\documentclass[sigconf]{acmart}
\AtBeginDocument{%
  }

\setcopyright{none}
\copyrightyear{2026}
\acmYear{2026}
\acmDOI{XXXXXXX.XXXXXXX}
\acmConference[WWW '26]{Make sure to enter the correct
  conference title from your rights confirmation email}{April 13--17,
  2026}{Dubai, United Arab Emirates}
\acmISBN{978-1-4503-XXXX-X/2018/06}

\usepackage{subcaption}
\usepackage{algorithm}
\usepackage{algorithmic}
\usepackage{multirow}
\usepackage{tabularx}

\usepackage{changes}

\definechangesauthor[name={Quan Zhou}, color=orange]{zq}

\begin{document}

\title{ViTs: Teaching Machines to See Time Series Anomalies Like Human Experts}

\author{Zexin Wang}
\authornote{Also with University of Chinese Academy of Sciences.}
\affiliation{%
  \institution{Computer Network Information Center, Chinese Academy of Sciences}
  \city{Beijing}
  \country{China}
}

\author{Changhua Pei}
\authornote{Corresponding author. Email: chpei@cnic.cn. Also with Hangzhou Institute for Advanced Study, University of Chinese Academy of Sciences.}
\author{Yang Liu}
\affiliation{%
  \institution{Computer Network Information Center, Chinese Academy of Sciences}
  \city{Beijing}
  \country{China}
}

\author{Hengyue Jiang}
\affiliation{%
  \institution{Hangzhou Institute for Advanced Study, University of Chinese Academy of Sciences}
  \city{Beijing}
  \country{China}
}

\author{Quan Zhou}
\author{Haotian Si}
\author{Hang Cui}
\affiliation{%
  \institution{Computer Network Information Center, Chinese Academy of Sciences}
  \city{Beijing}
  \country{China}
}

\author{Jianhui Li}
\authornote{Also with Nanjing University.}
\author{Gaogang Xie}
\author{Jingjing Li}
\affiliation{%
  \institution{Computer Network Information Center, Chinese Academy of Sciences}
  \city{Beijing}
  \country{China}
}

\author{Dan Pei}
\affiliation{%
  \institution{Tsinghua University}
  \city{Beijing}
  \country{China}
}

\renewcommand{\shortauthors}{Wang et al.}

\begin{abstract}

Web service administrators must ensure the stability of multiple systems by promptly detecting anomalies in Key Performance Indicators (KPIs). Achieving the goal of “train once, infer across scenarios” remains a fundamental challenge for time series anomaly detection models. Beyond improving zero-shot generalization, such models must also flexibly handle sequences of varying lengths during inference—ranging from one hour to one week—without retraining. Conventional approaches rely on sliding-window encoding and self-supervised learning, which restrict inference to fixed length inputs. Large Language Models (LLMs) have demonstrated remarkable zero-shot capabilities across general domains. However, when applied to time series data, they face inherent limitations due to context length. To address this issue, we propose \textbf{ViTs}, a Vision-Language Model (VLM)-based framework that converts time series curves into visual representations. By rescaling time series images, temporal dependencies are preserved while maintaining a consistent input size, thereby enabling efficient processing of arbitrarily long sequences without context constraints. Training VLMs for this purpose introduces unique challenges, primarily due to the scarcity of aligned time series image–text data. To overcome this, we employ an evolutionary algorithm to automatically generate thousands of high-quality image–text pairs and design a three-stage training pipeline consisting of: (1) time series knowledge injection, (2) anomaly detection enhancement, and (3) anomaly reasoning refinement. Extensive experiments demonstrate that ViTs substantially enhance the ability of VLMs to understand and detect anomalies in time series data. All datasets and code will be publicly released at: \url{https://anonymous.4open.science/r/ViTs-C484/}.

\end{abstract}





\maketitle

\section{Introduction}

Rapid detection of anomalies in Key Performance Indicators (KPIs) and timely recovery from failures are essential for ensuring the reliability of web systems \cite{fcvae,donut,pei2025flow}. However, with the continuous expansion of web services, the number of time series requiring monitoring has grown substantially. Consequently, enabling a “train once, infer across scenarios” paradigm with zero-shot capability has become a critical challenge for Time Series Anomaly Detection (TSAD) models. Moreover, KPIs exhibit considerable heterogeneity (e.g., periodicity), which results in significant variations in the optimal detection window size. An effective model should therefore be able to adaptively handle time series of different lengths during inference. Unfortunately, existing deep learning-based TSAD methods \cite{fcvae,donut,anomaly-transfomer,tuli2022tranad} are typically designed for fixed window sizes. Owing to their limited model capacity, these models must be re-trained or fine-tuned when applied to new scenarios, and thus lack the desired zero-shot generalization ability.

% \replaced{Having established the “train once, infer across scenarios” paradigm in the text domain, Large Language Models (LLMs) are now expected to demonstrate comparable success in time-series analysis, sparking growing research interest~\cite{nipsllmts1,nipsllmts2,nipllmts3,nipsllmts4}.}{With the rapid advancement of Large Language Models (LLMs), their potential in time series analysis has become increasingly evident \cite{nipsllmts1,nipsllmts2,nipllmts3,nipsllmts4}. Compared to traditional deep learning-based methods \cite{fcvae,anomaly-transfomer,nipstsad1,nipstsad2,si2023beyond}, LLMs, owing to their massive capacity, excel in interpretable zero-shot anomaly detection. Consequently, many} \added{Some} recent studies have attempted to directly feed time series data into LLMs in textual form, leveraging either prompt engineering \cite{llmad} or fine-tuning techniques \cite{xie2024chatts} to construct specialized Time Series LLMs (TS-LLM \added{in Figure \ref{fig:intro}}). However, representing time series as text introduces significant challenges: (i) extremely long context lengths, which hinder the inference of long time series, and (ii) unstable numerical token relationships, which often result in incoherent or inconsistent outputs.

\begin{figure*}[tbp]
    \centering
    \includegraphics[width=0.84\textwidth]{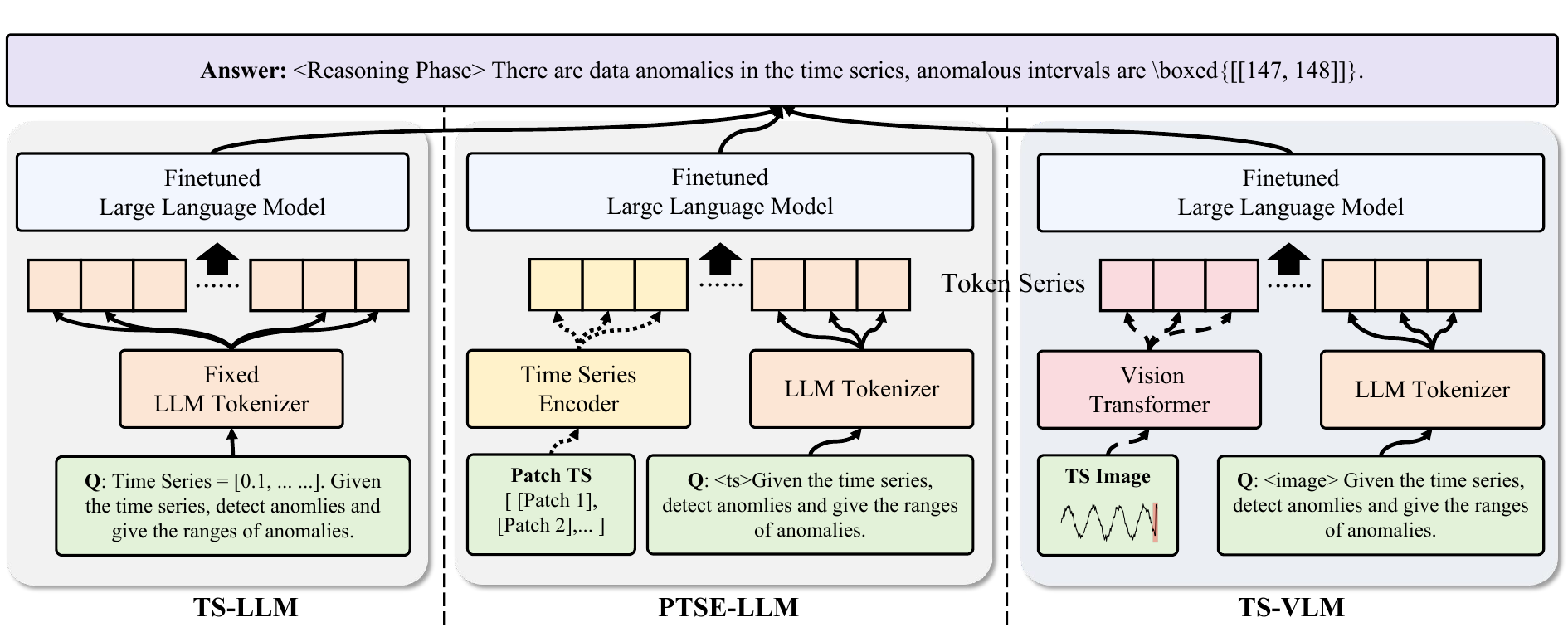} % 替换为你的文件名
    \caption{Comparison of TS-LLM, PTSE-LLM, and TS-VLM
.}
    \label{fig:intro}
\end{figure*}

Having established the “train once, infer across scenarios” paradigm in the text domain, Large Language Models (LLMs) are expected to demonstrate comparable success in time series analysis, sparking growing research interest~\cite{nipsllmts1,nipsllmts2,nipllmts3,nipsllmts4}. Some studies have attempted to directly feed time series data into LLMs in textual form, leveraging either prompt engineering \cite{llmad} or fine-tuning techniques \cite{xie2024chatts} to construct Time Series LLMs (TS-LLM in Figure \ref{fig:intro}). However, representing time series as text introduces significant challenges: (i) extremely long context lengths, which hinder the inference of long time series, and (ii) unstable numerical token relationships, which often result in incoherent or inconsistent outputs. To address the challenges, several studies~\cite{xie2024chatts} have proposed dedicated time series encoders that are jointly trained with textual modalities to achieve cross-modal alignment (Patch-based Time Series Encoders LLMs, PTSE-LLMs, in Figure~\ref{fig:intro}). These methods effectively mitigate the issue of context window overflow in LLMs when processing long time series. However, introducing a separate time series encoder inevitably imposes a fixed input length constraint, limiting the model’s ability to generalize to sequences longer than those seen during training. 

To address these problems, we draw inspiration from the analytical practices of human experts. We observe that, when determining whether a time series exhibits anomalies, experts typically do not inspect each data point in isolation as deep learning models often do. Instead, they visualize the time series curve on a dashboard and make holistic judgments by analyzing its overall patterns. Motivated by this insight, we propose a TSAD approach based on Vision Language Models (VLMs). By converting time series curves into images, our method effectively emulates the cognitive process of human experts. During downstream inference, time series of varying lengths are proportionally rescaled into fixed-size images, ensuring consistency with the training phase and maintaining stable inference performance. It is important to emphasize that time series image rescaling is fundamentally different from curve point sampling. Specifically, image rescaling only scales the indices of the time series points proportionally (e.g., $[0, 1, \ldots, 800] \rightarrow [0, 0.25, \ldots, 200]$) without altering the corresponding signal values. In contrast, direct downsampling of the raw time series modifies its intrinsic shape—sharp peaks, for instance, may become smoothed or flattened due to averaging over multiple points.

Although TSAD based on VLMs holds significant promise, directly applying existing general VLMs to time series analysis yields unsatisfactory results. The primary reason is that these models have not been pre-trained on large-scale datasets containing time-series–related images. In addition, they lack exposure to anomalous time series samples and corresponding analytical dual-modal text data, both of which are crucial for strengthening their capability in TSAD. To tackle these challenges, we propose \textbf{ViTs}, a TS-VLM specifically designed for TSAD tasks. To enhance data diversity, ViTs introduces a novel Seasonal-Trend decomposition using Loess (STL)-based approach \cite{stl} for periodic data generation and incorporates a wide range of anomaly injection strategies. In terms of training, we develop a three-stage chain-style fine-tuning framework, termed \textbf{Chain-of-TS}, which comprises: (1) \textit{time series knowledge injection}, (2) \textit{anomaly detection enhancement}, and (3) \textit{anomaly reasoning refinement}. 

% Moreover, to address the challenge of handling time series with varying lengths during inference, ViTs adopts a fixed length training and adaptive length inference mechanism.

% The main contributions of this work are as follows:
% \begin{enumerate}

%     \item 我们探索了不同time series large model 证明了 VLM的有效性，同时探索了不同画图方式对TS-VLM的影响。
%     \item We propose \textbf{ViTs}, a novel TS-VLM designed for TSAD. We introduce a new three-stage fine-tuning strategy called \textbf{Chain-of-TS}. By employing various types of data and diverse fine-tuning methods, this strategy fundamentally enhances the potential and effectiveness of VLMs in TSAD.
%     \item We propose a new fixed length training adaptive length inference schema to make ViTs fits for any time series length.
%     \item Experimental results show that ViTs outperforms both open-source and proprietary state-of-the-art (SOTA) VLMs by over 20\%, despite utilizing a relatively small number of parameters. All datasets and code are available at: \url{https://anonymous.4open.science/r/ViTs-C484/}.
% \end{enumerate}

The main contributions can be summarized as follows:
\begin{enumerate}

    \item We conduct a comprehensive exploration of different time series large models and demonstrate the effectiveness of VLMs. In addition, we investigate how various visualization approaches impact the performance of TS-VLMs. 

    \item We propose \textbf{ViTs}, a novel TS-VLM specifically designed for TSAD. To this end, we introduce a three-stage fine-tuning strategy, termed \textbf{Chain-of-TS}, which leverages diverse data types and fine-tuning techniques to substantially enhance the capability and effectiveness of VLMs in TSAD. 

    \item We design a new fixed length training and adaptive length inference paradigm. Time series image rescaling during inference enables ViTs to generalize effectively across time series of arbitrary lengths. 

    \item Extensive experiments show that ViTs surpasses state-of-the-art (SOTA) VLMs by more than 20\%, despite utilizing a relatively modest number of parameters. 
    % All datasets and code will be made publicly available at: \url{https://anonymous.4open.science/r/ViTs-C484/}. 

\end{enumerate}

\label{introduction}

\section{Preliminaries}
\subsection{Problem Definition}

Given a time series $X = \{x_1, x_2, \ldots, x_n\}$ collected at regular intervals, where $x_i$ denotes the value of a specific metric at time step $i$, the task of TSAD is defined as follows: 

At time $t$, a sliding window of length $w$ is extracted, denoted as $X_t = \{x_{t-w}, \ldots, x_t\}$. The goal is to assign an anomaly score to each point within the window, resulting in a score sequence $S_t = \{s_{t-w}, \ldots, s_t\}$. These scores are then used to identify anomalous intervals $L = \{[start_1, end_1], [start_2, end_2], \ldots\}$ via thresholding methods. Alternatively, the TSAD task can also be formulated as directly detecting anomalous intervals from the raw time series $X$ without explicitly computing point-wise anomaly scores.

\begin{table*}[tbp]
\small
\centering
\caption{The performance of different types of time series large models on TSAD. Both TS-LLM and TS-VLM utilize Qwen2.5-VL-7B-Instruct. P indicates precision, R indicates recall, and F1 represents the best F1 score after point adjustment. The best results are highlighted in \textbf{bold}, and the second best are \underline{underlined}.}
\label{tab:ts-llm}
\begin{tabularx}{\textwidth}{l *{12}{>{\centering\arraybackslash}X}} 
\toprule
                       &   &    \textbf{Spike}         &            &  & \textbf{Trend}            &            & &      \textbf{Frequency}       &            &  &  \textbf{Level}            \\ 
\cmidrule{2-13}
                       & \textbf{P} & \textbf{R}     & \textbf{F1} & \textbf{P} & \textbf{R}     & \textbf{F1} & \textbf{P} & \textbf{R}         & \textbf{F1} & \textbf{P} & \textbf{R}           & \textbf{F1}  \\ 
\midrule
\textbf{TS-LLM}        & 0.4496     & 0.0099         & 0.0194      & 0.3327     & 0.0316         & 0.0577      & 0.0000     & 0.0000             & 0.0000      & 0.1660     & 0.0086               & 0.0164       \\
\textbf{TS-VLM}        & 0.5631     & 0.5490         & 0.5560      & 0.7019     & 0.0913         & \underline{0.1616}      & 0.8993     & 0.3035             & 0.4538      & 0.5733     & 0.5819               & 0.5776       \\
\textbf{PTSE-LLM(SFT)} & 0.1821     & 0.1432         & 0.1605      & 0.1954     & 0.1025         & 0.1348      & 0.1732     & 0.1520             & 0.1620      & 0.1610     & 0.1245               & 0.1406       \\
\textbf{TS-LLM(SFT)}   & 0.9434     & 0.7385         & \underline{0.8286} & 1.0000        & 0.0326         & 0.0631      & 0.9964     & 0.3840             & \underline{0.5543}      & 0.9465     & 0.8155               & \underline{0.8761}       \\
\textbf{TS-VLM(SFT)}   & 0.9347     & 0.8203         & \textbf{0.8737} & 0.9365     & 0.1543         & \textbf{0.2649} & 0.9328     & 0.4813             & \textbf{0.6350} & 0.9460     & 0.8732               & \textbf{0.9081} \\
\bottomrule
\end{tabularx}
\end{table*}

\subsection{Anomaly Types}

Based on extensive observations of public time series datasets, this paper focuses primarily on periodic time series and categorizes time series anomalies into four types: spike anomaly, trend anomaly, level anomaly, and frequency anomaly. Details can be find in Section \ref{sec:ano} and Figure \ref{fig:3x4grid}.

\label{preliminaries}

\section{Motivation}

\subsection{TS-LLM \& TS-VLM \& PTSE-LLM}

% 在Section \ref{introduction}中，我们介绍了现在的时间序列大模型的构建方式大概有三种，分别是TS-LLM TS-VLM和PTSE-LLM。为了验证，TS-VLM是一种更简单有效的方式我们进行了实验。具体来说，针对TS-LLM TS-VLM 我们分别比较了不同方式在微调前以及SFT微调之后 TSAD的效果。这里不同方法使用的微调数据相同（10k 个 人造QA数据集），评估数据使用2k个人造QA数据（与Section \ref{evaluation}中评估人造评估数据集相同），使用的模型为Qwen2.5-VL-7B-Instruct(即TS-LLM使用只使用其LLM Backbone， TS-VLM同时使用Vision Encoder)。针对PTSE-LLM，我们直接使用微调后的ChatTS。微调前后的结果如表\ref{}所示。

In Section \ref{introduction}, we introduced three main approaches for constructing time series large models: TS-LLM, TS-VLM, and PTSE-LLM. To verify that TS-VLM is a simpler and more effective approach, we conducted experiments. Specifically, we compared the performance of TS-LLM and TS-VLM both before and after STF with respect to TSAD. The same fine-tuning data (10k synthetic QA datasets) was used for different methods, with evaluation performed using 2k synthetic QA datasets (consistent with the synthetic evaluation datasets described in Section \ref{evaluation}). The models used were Qwen2.5-VL-7B-Instruct (with TS-LLM utilizing only its LLM backbone, and TS-VLM utilizing both the Vision Encoder and LLM backbone). For PTSE-LLM, we directly used the fine-tuned ChatTS. The results before and after fine-tuning are shown in Table \ref{tab:ts-llm}.

% %可以看出在SFT 微调之前，TS-LLM几乎没有TSAD能力。这是因为输入的文本形式的时序会产生大量的数字token，本身LLM并没有专门在序列话的数字token上进行针对性训练，因此LLM很容易会产生不断复读数字、回答和问题不对应等情况，导致TSAD效果很差。而TS-VLM则存在很大不同，即便是naive的没有经过微调的Qwen2.5-VL-7B-Instruct已经能够正确检测出很大一部分异常，这是因为，不同类型的图片之间存在共性，通用图片理解能力的训练也会增强时序图片的理解能力。而在SFT微调之后，可以发现，PTSE-LLM效果最差，这是可以理解的，因为从头训练一个time series encoder确实很困难。TS-LLM和TS-VLM微调之后效果均不错，TS-LLM由于会将原始值的大小输入，因此在spike和level shift这种值快速变化的异常上表现更好。而TS-VLM能够通过图片全面感知时序的波动周期与趋势，因此在trend和frequecny类型的异常上表现效果更好。

% 然而，如表\ref{}所示，随着时序长度的增大，TS-LLM的token消耗会显著增加，因此使用TS-VLM是一种更为简单有效的方式。

It can be observed that before SFT, TS-LLM has almost no TSAD capability. This is because the time series in text format generates a large number of numeric tokens, and LLMs are not specifically trained on sequential numeric tokens. Consequently, LLMs are prone to issues such as repetitive output of numbers and mismatches between answers and questions, leading to poor TSAD performance. On the other hand, TS-VLM shows significant differences. Even a naive, non-fine-tuned Qwen2.5-VL-7B-Instruct can correctly detect a substantial portion of anomalies. This is because there are commonalities among different types of images, and training in general image understanding enhances the ability to understand time series images. 

% After SFT, it becomes apparent that PTSE-LLM performs the worst, which is understandable because training a time series encoder from scratch is quite challenging. Both TS-LLM and TS-VLM perform well after fine-tuning. TS-LLM performs better on anomalies like spikes and level shifts, which involve rapid changes in values, because it inputs the raw values. In contrast, TS-VLM effectively perceives the periodic and trend fluctuations of the time series through images, hence it performs better on trend and frequency anomalies. However, as shown in Figure \ref{}, the token consumption of TS-LLM significantly increases with the augmentation of time series length, making TS-VLM a simpler and more effective approach.
% After SFT, it becomes apparent that PTSE-LLM performs the worst, which is understandable because training a time series encoder from scratch is quite challenging. TS-VLM则在各类异常上均超过了TS-LLM。这是很容易理解的。通用图片的理解能力可以帮助VLM更好的理解时序图片，而文本形式的time series 很难理解，并且会消耗大量的token。 Therefore, using TS-VLM is a more suitable approach.
After SFT, it becomes apparent that PTSE-LLM performs the worst, which is understandable because training a time series encoder from scratch is quite challenging. TS-VLM outperforms TS-LLM across various anomaly types. This is easy to understand. The general image understanding capabilities of VLMs can significantly enhance their ability to interpret time series images, whereas time series in text form are difficult to understand and consume a large number of tokens. Therefore, using TS-VLM is a more suitable approach.

\subsection{Comparison of Plot Type}

% 对于通用任务包括TSAD，我们给予LLM越详细的信息，其完成的准确率大概率会越高。那么在VLM领域是否也是如此？为了验证这个猜想是否正确，我们进行了实验。由于VLM包括vision 模态和语言模态，而语言模态利用的是目前已经有的LLM backbone，因此我们这里的实验主要验证给予更多的vision 模态的信息是否会提高TSAD的效果。
% 首先面临的问题是我们输入VLM什么信息。通用的时序画图方法是时序曲线图，简介直观的表示时序的波动性，这也是大多数方法所采用的。然而，现在越来越多的方法证明频域信息的重要性，而VLM本身并不具备从曲线图提取频域信息的能力。因此我们验证输入频域图这种额外只是是否会对于TSAD效果有提升。这里频域图的种类我们采用了目前通用的STFT和Wavelet。
% 其次是如何输入VLM vision模态额外的信息。这里我们尝试了两种方法。一种是直接输入VLM包括曲线图和频域图在内的多张图片。另一种方式则是将曲线图和频域图当作两张子图集成到一张图片上。
%实验结果如图1所示。可以看出，在不SFT的基础上，分开输入多张图片并没有提升f1 score，而将频域图和线性图拼接为一张图片效果却提升明显。这是因为我们将频域图和线性图在同一张子图上进行了index对齐。某些异常例如频域异常在线性图上很难看出，但是转换为频域图很容易观测到。
%为了进一步提升TSAD的效果，我们用大量人造数据进行SFT，然而SFT之后的结果却并并不好。线性图的效果反而是最好的。为了探究原因，我们使用Qwen2.5-VL-7B-Instruct分别描述不同类型的图片，结果如图2所示。我们发现，当输入一张线性图片时，VLM会针对图片的重要细节进行描述。然而，当输入两个子图构成的图片时，VLM会更多的描述一些不重要的信息，例如每张图片的x轴、y轴是什么，这和TSAD相关性不大。

% 因此，即便多张图片多种类型的输入可能会有提升，但是其上限却相对较低，因此本篇paper仍旧使用线性图一张图片进行输入。

For general tasks, providing LLMs with more detailed information typically enhances their performance \cite{cot,shinn2023reflexion}. We hypothesize that this principle might also apply in VLMs. To test this assumption, we conducted a series of experiments. Considering that VLMs comprise both vision and textual modalities—and that the textual modality generally relies on pre-trained LLM backbones—our investigation centers on whether more detailed visual input can enhance TSAD performance.

The initial challenge is determining what kind of visual information should be fed into the VLM. The most common visualization method for time series data is the line chart, which intuitively illustrates temporal fluctuations. While widely adopted, recent research increasingly underscores the importance of frequency-domain information \cite{fcvae,freq1,xu2023fits}. However, VLMs do not inherently extract such information from line charts. Therefore, we examine whether augmenting the input with frequency-domain visualizations—specifically Short-Time Fourier Transform (STFT) \cite{stft} and Wavelet Transform \cite{zhang2019wavelet} plots—can improve TSAD performance (shown in Figure \ref{fig:3x4grid}). The next question concerns how to incorporate this additional visual information. We explored two strategies: (1) Supplying multiple images to the VLM, including both line charts and frequency-domain plots; and (2) Merging the line chart with the frequency-domain plots into a single composite image with subplots. (Examples shown in Section \ref{sec:visual}.)

\begin{figure}
  \centering
  \includegraphics[width=0.48\textwidth]{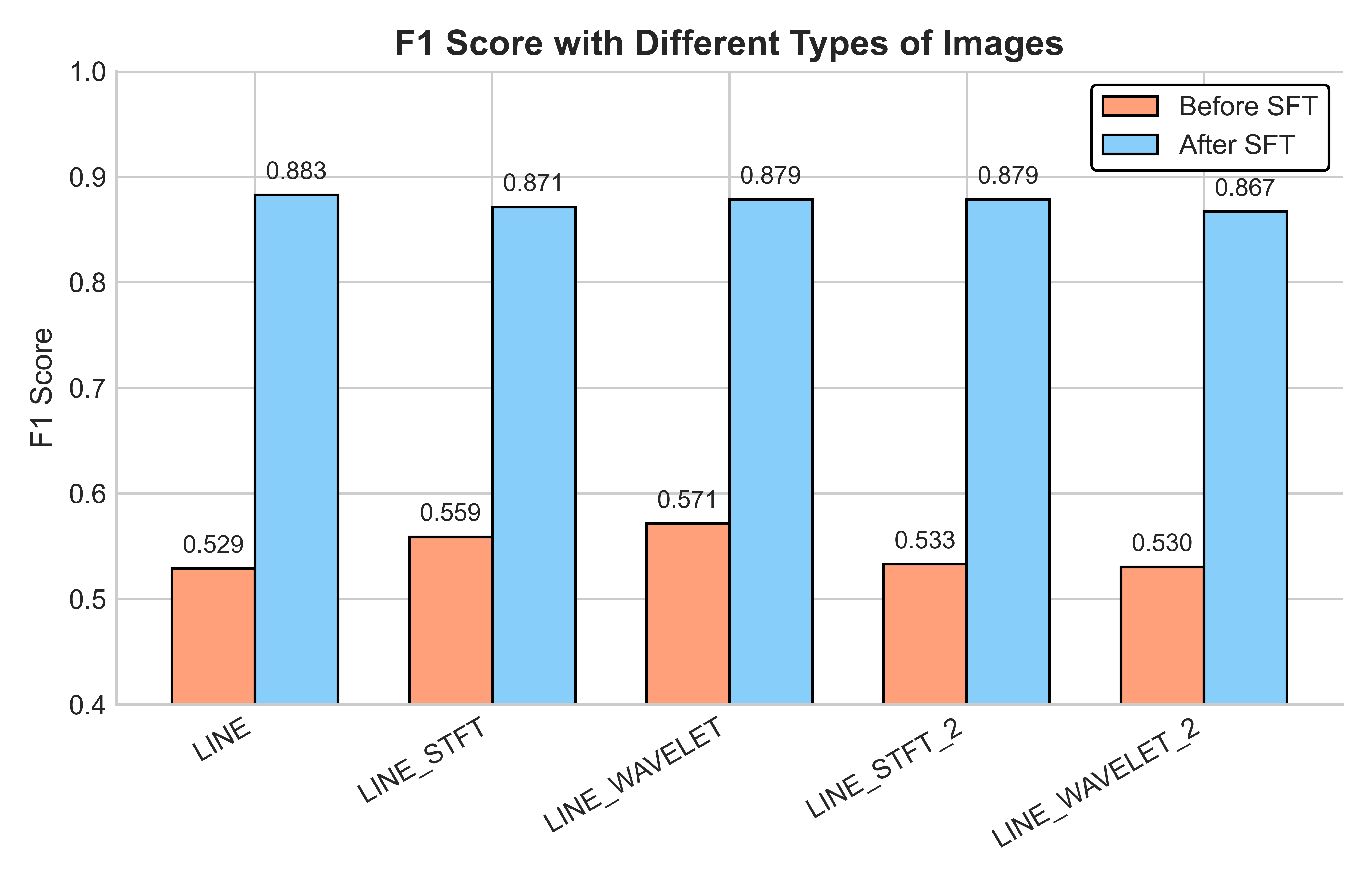} % 替换为你的图片名
  \caption{Performance across different image types. \textbf{LINE} refers to a single line plot. \textbf{LINE\_STFT} denotes a composite image containing two subplots—one showing the line plot and the other displaying the STFT. \textbf{LINE\_STFT\_2} indicates two separate input images: one for the line plot and another for the STFT.}
  \label{fig:motivation}
\end{figure}

The experimental results, depicted in Figure \ref{fig:motivation}, demonstrate that supplying multiple separate images does not improve the F1 score without SFT. In contrast, combining the line and frequency-domain plots into a single image results in a noticeable performance improvement. We attribute this to the index alignment across subplots, which enables the model to correlate temporal and frequency-domain anomalies more effectively. Certain anomalies, which may be subtle in the time domain, become more apparent in the frequency domain.

To further enhance TSAD performance, we applied SFT using a large volume of synthetic data. However, contrary to expectations, the results deteriorated, with line charts alone delivering the best performance. To investigate this, we employed Qwen2.5-VL-7B-Instruct to generate image descriptions for various input formats (results shown in Section \ref{imagetype}). We observed that when exposed to a single line chart, the VLM produced detailed and relevant descriptions. However, for composite images with subplots, the VLM focused more on superficial elements, such as axis labels, which are not directly pertinent to TSAD.

% \noindent\fbox{%
%   \parbox{\linewidth}{%
%   \small
%     \textbf{LINE:} \textit{Key observations from the chart:
%     1. The series starts at a low value around 0.00 and gradually increases. 
%     2. There are fluctuations in the series, indicating variability or noise in the data. 
%     3. A significant peak occurs around the 125th index, reaching a value close to 1.00. 
%     4. After the peak, the series fluctuates but generally decreases and stabilizes around a value of 0.50.}
%     \\
%     \textbf{LINE\_STFT:} \textit{The image consists of two plots. The top plot is labeled "Time Series (Index-based)" and shows a time series of data indexed from 0 to 160. The y-axis represents the "Value," which ranges from 0 to 1. The data appears to be a noisy signal with some peaks and troughs, particularly noticeable around index 125 where there is a sharp peak. The bottom plot is labeled "STFT Magnitude Spectrum."} 
%   }%
% }

Therefore, although using multiple image types may offer some marginal gains, the potential improvement appears limited. As such, this paper adopts single line charts as the visual input.

\label{motivation}

\section{ViTs}

% ViTs introduces a novel STL-based periodic time series generator, an Attribute-based Time Series Description Generator,along with a new Chain-of-TS fine-tuning strategy tailored for TSAD. The overall workflow is illustrated in Figure \ref{fig:model}.
% ViTs introduces a novel STL-based periodic time series generator and an attribute-based time series description generator, along with a new Chain-of-TS fine-tuning strategy tailored for TSAD. The overall workflow is illustrated in Figure \ref{fig:model}.

\begin{figure*}[tbp]
    \centering
    \includegraphics[width=0.84\textwidth]{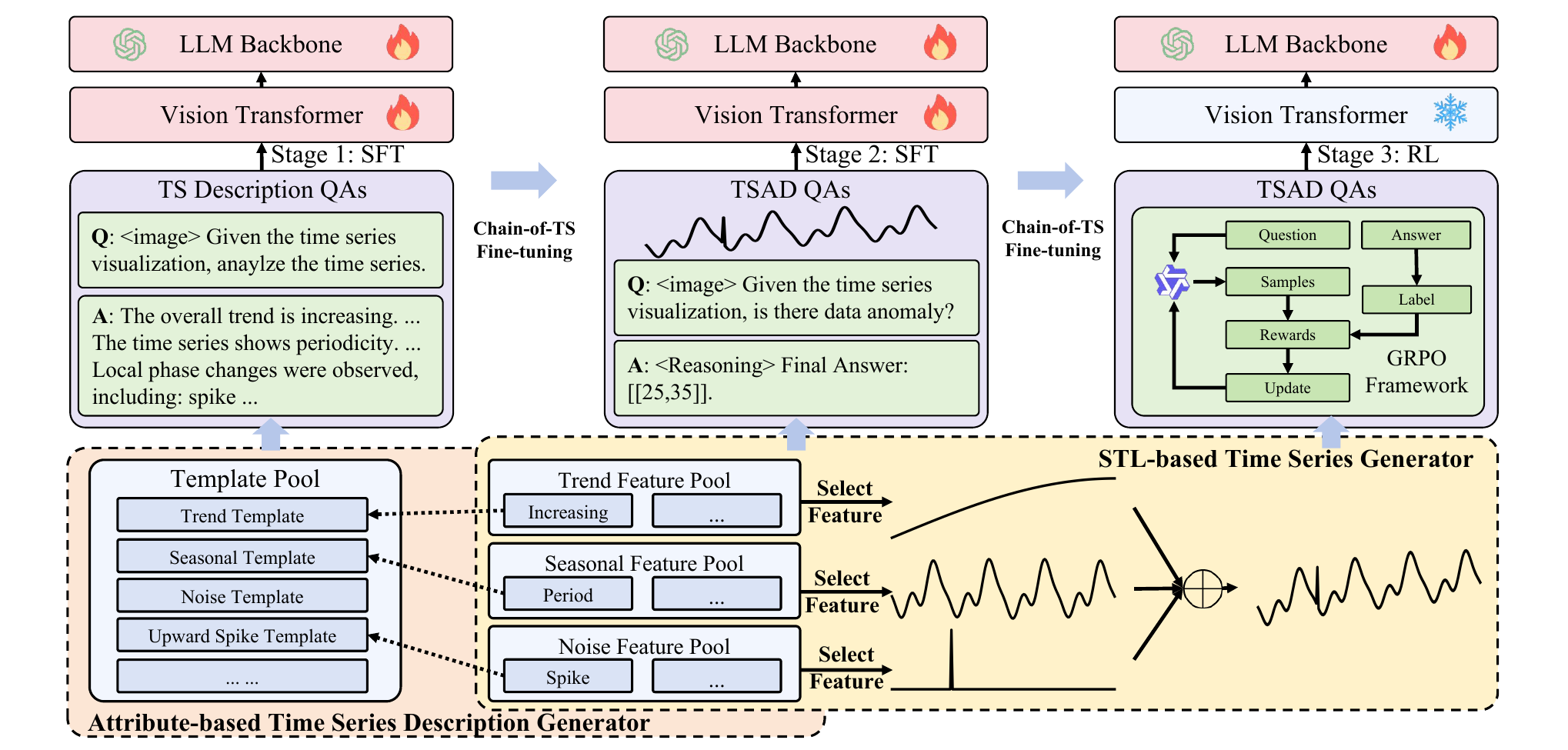} % 替换为你的文件名
    \caption{Overview of ViTs.}
    \label{fig:model}
\end{figure*}

The overall workflow of ViTs is illustrated in Figure \ref{fig:model}.

\subsection{STL-based Time Series Generator}

To enhance the zero-shot TSAD capabilities of VLMs, it is crucial to train them with a substantial volume of time series. According to the STL theory, any time series can be decomposed into three components: seasonal, trend, and noise. Based on this principle, we propose an STL-based time series generator that synthesizes time series by generating these three components and then combining them. 

\subsubsection{Trend}

%生成trend时候，我们随机选择increase、decrease 和keep steady中的三种。在increase和decrease 中，我们设计了多种升降模式，其中包括sin、指数、线性等。trend的强度也随机确定。

When generating trends, we randomly choose from three types: increase, decrease, and keep steady. For increase and decrease, we design multiple modes, including exponential, linear, and others. The intensity of the trend is determined randomly.

\subsubsection{Seasonal}

For the seasonal component, Fourier series theory \cite{fourier1888theorie} demonstrates that any periodic function $f(x)$ can be represented as an infinite sum of sine and cosine functions. Drawing on insights from KAN-AD \cite{kanad}, which shows that most normal time series patterns can be well-approximated using only a small number of sine and cosine terms $S_N(x)$ (proof detail in Section \ref{proof}), we generate the seasonal signal by randomly combining a limited set of such functions.
{\small
\[
f(x) \;=\; \frac{a_0}{2} + \sum_{n=1}^{\infty} \bigl(a_n \cos nx + b_n \sin nx\bigr)
\]
\[
S_N(x) \;=\; \frac{a_0}{2} + \sum_{n=1}^N \bigl(a_n \cos nx + b_n \sin nx\bigr), \bigl\lvert f(x) - S_N(x)\bigr\rvert \;\le\; \sigma
\]
}

% \begin{algorithm}
% \caption{Seasonal Generation}
% \begin{algorithmic}[1]
% \REQUIRE period_type, amplitude_series
% \STATE Randomly choose whether to use a large period or a small period
% \IF{period_type is long}
%     \STATE $period \gets random(10, \lfloor ts\_length / 2 \rfloor)$
% \ELSE
%     \STATE $period \gets random(\lfloor ts\_length / 2 \rfloor, 3 \times ts\_length)$
% \ENDIF

% \STATE $base\_frequency \gets 1 / period$
% \STATE $num\_harmonics \gets random\_integer(1, 10)$

% \FOR{$n = 1$ to $num\_harmonics$}
%     \STATE $phase \gets random\_uniform(0, 2\pi)$
%     \STATE $harmonic\_amplitude \gets \dfrac{amplitude\_series}{n} \times \Big(1 + random\_uniform(0, 0.05) \cdot \sin\big(random\_uniform(1, 3)\pi \cdot t / ts\_length + random\_uniform(0, 2\pi)\big)\Big)$
%     \STATE $data \gets data + harmonic\_amplitude \cdot \sin(2\pi \cdot base\_frequency \cdot n \cdot t + phase)$
% \ENDFOR
% \end{algorithmic}
% \label{algothim}
% \end{algorithm}

\begin{algorithm}[t]
\caption{Seasonal Generation}
\label{alg:seasonal_generation}
\begin{algorithmic}[1]
\REQUIRE $period\_type$, $amplitude\_series$, $ts\_length$
\ENSURE Generated seasonal data $data$

\STATE Initialize $data \gets 0$

\IF{$period\_type$ is long}
    \STATE $period \gets random(10, \lfloor ts\_length / 2 \rfloor)$
\ELSE
    \STATE $period \gets random(\lfloor ts\_length / 2 \rfloor, \, 3 \times ts\_length)$
\ENDIF

\STATE $base\_frequency \gets 1 / period$
\STATE $num\_harmonics \gets random\_integer(1, 10)$

\FOR{$n = 1$ to $num\_harmonics$}
    \STATE $phase \gets random\_uniform(0, 2\pi)$
    \STATE $harmonic\_amplitude \gets \dfrac{amplitude\_series}{n} 
           \times \Big(1 + random\_uniform(0, 0.05) \cdot 
           \sin\big(random\_uniform(1, 3)\pi \cdot t / ts\_length 
           + random\_uniform(0, 2\pi)\big)\Big)$
    \STATE $data \gets data + harmonic\_amplitude \cdot 
           \sin(2\pi \cdot base\_frequency \cdot n \cdot t + phase)$
\ENDFOR
\end{algorithmic}
\end{algorithm}

%具体来说，如算法\ref{algo}所示，我们首先随机选取周期长度，这里我们着重区分large period 和 small period，主要是为了后续的周期描述信息生成过程使用。因为如果说时序的周期小于time series length的一半，那么大概率观测图像无法看出显著的周期性。并且在真实数据中，经常会出现一个窗口无法包含周期的情况。之后根据周期确定基础频率，并随机选取num_harmonics个频率分量，生成频域分量对应的时序数据并相加。需要注意的是，不同频域分量对应的强度是不同的，一般来说低频分量强度更大，高频分量（大概率是噪声）强度更小。因此我们用1/n控制高频分量的强度，同时引入随机性保证幅度的随机扰动。通过这种方式，我们生成了大量随机的周期分量。

Specifically, as shown in Algorithm \ref{alg:seasonal_generation}, we first randomly select a period length, distinguishing between large periods and small periods to facilitate the subsequent generation of periodic description information. This distinction is essential because if the period of the time series is less than half of the time series length, it is unlikely that significant periodicity will be observable in the image. In real-world data, it is common for a window not to contain a full period. Next, we determine the fundamental frequency based on the period and randomly select $num\_harmonics$ frequency components. We then generate time series data corresponding to these frequency components and sum them up. It is important to note that different frequency components have different intensities. Generally, low-frequency components have higher intensities, while high-frequency components (which are likely noise) have lower intensities. Therefore, we use 1/n to control the intensity of high-frequency components and introduce randomness to ensure random perturbations in amplitude. With this approach, we generate a large number of random periodic components.

\subsubsection{Noise}

% 在noise 插入阶段，我们根据整个时序的amplitude来随机选择high 或者low noise。更重要的是，在这里我们在时序中插入异常。为了保证异常插入的种类足够diverse，我们分析了大量真实数据总结\ref{preliminaries}中的四类异常。每类异常还包含多种实现方式。如表\ref{ano}所示。Spike是真实世界时序数据中最常见的一类异常，但是spike的模式多种多样，因此我们将其区分为upward spike、downward spike等多种形式。Level shift同样是真实世界常见的异常，经常发生在系统变更时。然而 除了常见的 sudden increase、decrease之外，在变更时可能突然出现一个spike之后才发生level shift，因此我们同样引人decrease after upward spike等。对于trend形式的异常，我们直接随机选取一段时序随机生成一段新的trend来插入异常。

% 对于frequency形式的异常，我们采用如\ref{}所示的方式，插入异常。具体来说，首先随机选择插入区域，之后将这部分时序转为频域，在频域上施加一定的扰动，再转换回时域。这样，这部分时序的周期性和周围的正常时序存在明显不同。需要注意的是，我们选择扰动频域分量的时候，尽可能随机选择强度比较大的频域分量，因为较小的频域分量可能是噪声，即便是扰动了也可能不会对于时序周期性产生显著影响。

During the noise insertion phase, we randomly select high or low noise based on the amplitude of the entire time series. More importantly, we insert anomalies into the time series. To ensure a sufficient diversity of anomaly types, we analyzed a large amount of real data and summarized the four types of anomalies in Section \ref{preliminaries}. Each type of anomaly includes multiple implementation methods, as shown in Table \ref{tab:ano}. 

\begin{figure}[tbp]
    \centering
    \includegraphics[width=0.39\textwidth]{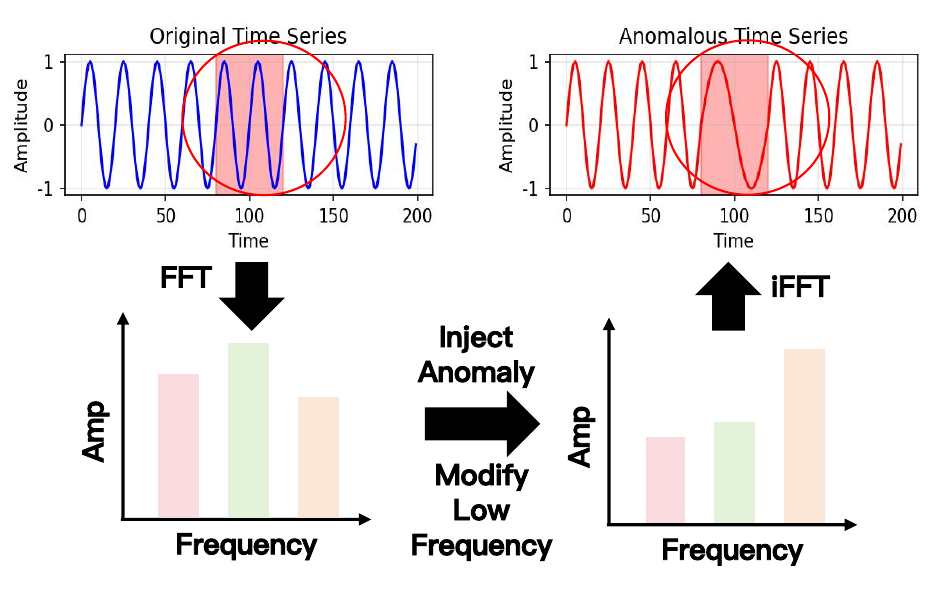} % 替换为你的文件名
    \caption{Illustration of frequency anomalies.}
    \label{fig:freqano}
\end{figure}

Spike is the most common type of anomaly in real-world time series data, but spike patterns vary widely. Therefore, we categorize them into multiple forms such as upward spike. Level shift is another common anomaly in real-world systems and frequently occurs during system changes. Besides the common sudden increases and decreases, a level shift might be preceded by a spike, resulting in types like decrease after an upward spike. For trend anomalies, we randomly select a segment of the time series and generate a new trend to be inserted as the anomaly. For frequency anomalies, as depicted in Figure \ref{fig:freqano}, we insert anomalies by modify the frequency domain. This ensures that the periodicity of this segment is significantly different from the normal time series around it. It is important to note that when selecting the frequency components to perturb, we randomly choose components with relatively high intensity, as smaller frequency components might be noise.

\begin{table}[tbp]
\centering
\small
\caption{Details of different types of anomalies.}
\label{tab:ano}
\begin{tabular}{p{1.3cm} p{6cm}} % 第一列3cm，第二列12cm，可根据需要调整
\toprule
\textbf{Type}       & \textbf{Category or Description} \\
\midrule
\textbf{Spike}       & Upward spike, Downward spike, Continuous upward spike, Continuous downward spike, Upward convex, Downward convex, Rapid rise followed by slow decline, Slow rise followed by rapid decline ...\\
% , Rapid decline followed by slow rise, Slow decline followed by rapid rise \\
\midrule
\textbf{Level} & Sudden increase, Sudden decrease, Increase after downward spike, Decrease after upward spike, Increase after upward spike ... \\
\midrule
\textbf{Trend}       & Random Generate Multiple Trend \\
\midrule
\textbf{Frequency}   & Modify Low Frequency Components \\
\bottomrule
\end{tabular}
\end{table}

\subsection{Attribute-based Time Series Description Generator}

% We demonstrate later that the primary reason for the suboptimal performance of current VLMs on TSAD lies in the vision encoder's inability to extract meaningful information relevant to TSAD—specifically, periodic and trend-related information. CLIP \cite{clip} has shown that the most effective way to train a vision encoder is through large-scale image–text pair datasets. Therefore, to enhance the vision encoder's ability to capture key features of time series, we propose an attribute-based time series description generator来生成时序描述相关的QA。具体来说，我们会根据之前STL阶段生成time series时所选择的attribute生成描述信息。描述信息主要包含四个方面，第一个时时序的trend，increase还是decrease等。第二个是时序的周期性，如果peried 小于时序length的一半，则认为有明显的周期性。第三部分是时序的局部异常，这包括我们插入的各类异常的描述。我们会针对每一类异常设置一个描述模版，并根据随机选择相关特征生成描述信息。第四部分是噪声强度，有的时序噪声很强，有的很弱。Examples and details can be found in Section \ref{sec:QAexample}.

% 总的来说，这部分时序描述QA从不同角度说明了时序的特征，学习这些特征对于后续的时序任务有着重要意义。

We demonstrate later that the primary reason for the suboptimal performance of current VLMs on TSAD lies in the vision encoder's inability to extract meaningful information relevant to TSAD—specifically, periodic and trend-related information. CLIP \cite{clip} has shown that the most effective way to train a vision encoder is through large-scale image–text pair datasets. Therefore, to enhance the vision encoder's ability to capture key features of time series, we propose an attribute-based time series description generator to generate time series-related QA. Specifically, we generate descriptive information based on the attributes selected during the previous STL stage when creating the time series. The descriptions mainly encompass four aspects:
(1) The trend of the time series: whether it is increasing, decreasing, etc.
(2) The periodicity of the time series: if the period is less than half the time series length, it is considered to have significant periodicity.
(3) Local anomalies in the time series: descriptions of various inserted anomalies. We set a description template for each type of anomaly and generate descriptive information.
(4) Noise intensity: some time series have strong noise, while others have weak noise.
Examples and details can be found in Section \ref{sec:QAExample}.

% In summary, these time series descriptive QAs explain the characteristics of time series from different angles, and learning these features is of significant importance for subsequent time series tasks.

\subsection{Chain-of-TS Fine-tuning Strategy}

% As illustrated in Figure \ref{fig:model}, Chain-of-TS involves a three-stage fine-tuning process. In the first stage, the vision encoder of the VLM is fine-tuned using SFT on time series description QAs. In the second stage, SFT is applied to the entire VLM using TSAD QAs, thereby aligning the visual and textual modalities. In the final stage, RL is employed to fine-tune the LLM backbone, allowing the model to continuously explore and enhance its TSAD capabilities. The following sections detail each stage.

As illustrated in Figure \ref{fig:model}, Chain-of-TS involves a three-stage fine-tuning process. 
% In the first stage, the vision encoder of the VLM is fine-tuned using SFT on time series description QAs. In the second stage, SFT is applied to the entire VLM using TSAD QAs, thereby enhancing the VLM's ability to utilize foundational knowledge to complete TSAD tasks and training its instruction-following capability. In the final stage, RL is employed to fine-tune the LLM backbone, allowing the model to continuously explore and enhance its TSAD capabilities. The following sections detail each stage.

% \begin{figure*}[tbp]
%     \centering
%     \includegraphics[width=0.9\textwidth]{pic/model_new.pdf} % 替换为你的文件名
%     \caption{Overview of ViTs.}
%     \label{fig:model}
% \end{figure*}

\subsubsection{Stage 1: Time Series Knowledge Injection with SFT}
% 在motivation章节中我们指出，限制TSAD效果的主要因素在于vision encoder 无法有效的捕捉与TSAD相关的细节信息，尤其是和周期性和trend相关的信息。因此，提升TSAD效果的关键在于提升vision encoder特征提取的能力。常用的微调vision encoder 的方式为用图像文本QA对利用CLIP算法微调。在此，我们为了方便，使用类似的方式达到等价于CLIP的效果。即，直接用 time series 和 描述信息组成的QA对，端到端的方式用SFT微调VLM。微调的过程中，只开放vision encoder的相关参数，冻结LLM backbone的参数。这里我们的数据采用的是上文attribute-based ts description generator生成的人造数据，其中包含了大量的周期性、trend等信息。后续的实验结果也证明了改阶段的重要作用，其在本质上提升了VLM对于时序理解的上限。

% In Section \ref{motivation}, we highlight that a key limitation in TSAD performance lies in the vision encoder's inability to effectively capture detail-rich features relevant to TSAD. Therefore, enhancing the feature extraction capabilities of the vision encoder is crucial for improving TSAD performance. A common approach for fine-tuning vision encoders involves using image-text QA pairs with the CLIP \cite{clip}. In our work, we adopt a similar strategy to achieve CLIP-equivalent performance by directly fine-tuning the VLM in an end-to-end manner. During this fine-tuning phase, only the vision encoder parameters are updated, while the parameters of the LLM backbone remain frozen. The training data used in this stage is synthetically generated by the previously introduced attribute-based time series description generator, which provides rich representations of periodicity, trends, and other relevant features. Subsequent experiments validate the significance of this stage, demonstrating that it substantially enhances the upper bound of the VLM’s ability to understand time series.

In Section \ref{motivation}, we highlight that a key limitation in TSAD performance lies in the vision encoder's inability to effectively capture detail-rich features relevant to TSAD. Therefore, enhancing the feature extraction capabilities of the vision encoder is crucial for improving TSAD performance. A common approach for fine-tuning vision encoders involves using image-text QA pairs with the CLIP \cite{clip}. In our work, we adopt a similar strategy to achieve CLIP-equivalent performance by directly fine-tuning the VLM in an end-to-end manner. During this fine-tuning phase, we open up all parameters, including the vision encoder. The training data used in this stage is QAs generated by the attribute-based time series description generator, which provides rich information on periodicity, trends, and other relevant features. Subsequent experiments validate the significance of this stage, demonstrating that it substantially enhances the upper bound of the VLM’s ability to understand time series.

\subsubsection{Stage 2: Anomaly Detection Enhancement with SFT}

% In the previous stage, we fine-tuned only the vision encoder of the VLM while keeping the LLM backbone frozen. Although this strategy effectively enhanced the vision encoder’s ability to extract time series features, it also introduced a misalignment between the visual and textual modalities. Addressing this misalignment is essential for effective multimodal understanding. Therefore, in the second stage, we unfreeze all parameters of the VLM and perform end-to-end fine-tuning using SFT on TSAD-related QA pairs. These QA pairs are generated using the previously described STL-based time series generator. 
In the first stage, we only taught the VLM how to understand time series images and how to better extract features from them. However, it still does not know how to perform time series tasks such as TSAD. Therefore, in the second stage, we open up all parameters and use TSAD-related QA to teach the TS-VLM how to efficiently and accurately complete TSAD tasks. Since we know the ground truth of the inserted anomalies, it is easy to obtain TSAD QA.

\subsubsection{Stage 3: Anomaly Reasoning Refinement with RL}

%通过上述两个阶段，我们从本质上提升了VLM在TSAD上的能力。然而SFT更多的是训练模式匹配的能力，存在部分过拟合的情况，模型内部本身的潜力并没有被完全释放出来。近些年的研究表明，RL可以通过采样的方式，在答案空间不断探索，并判断最优答案。这种方式能从本质上挖掘模型的潜力并释放，并且有更好的泛化性。因此，这里我们使用RL中的经典框架GRPO来挖掘VLM TSAD的潜力。GPPO中我们设置采样为5，并通过如下方式计算reward：

%这里我们直接使用f1_reward的原因是简单直接，format_reward则是使得模型能够按照规定的方式输出答案，同时提升模型reasoning 的能力。我们从训练过程中的reward曲线可以看出，少量的训练步骤变有效的提升了VLM TSAD的能力，证明了RL 后训练的有效性。

% \begin{wrapfigure}{r}{0.3\textwidth}
%   \centering
%   \includegraphics[width=0.28\textwidth]{pic/rl_reward.png} % 替换为你的图片名
%   \caption{Reward score of different training steps.}
%   \label{fig:reward}
% \end{wrapfigure}

Through the first two stages, we have significantly enhanced the VLM's capabilities for TSAD. However, SFT primarily encourages pattern matching and may lead to partial overfitting, leaving the model's full potential underexplored. Recent research suggests that RL can further unlock model capacity by enabling exploration within the answer space through sampling and iterative reward-based optimization. This approach not only helps uncover deeper capabilities of the model but also tends to offer improved generalization. To this end, we adopt the Group Relative Policy Optimization \cite{guo2025deepseek}(GPPO, detailed in Section \ref{sec:GRPO} and Figure \ref{fig:model}) to further explore and enhance the TSAD capabilities of the VLM.

% In our implementation, we set the number of samples to 5 and define the reward function as follows: 

% {
% \small
% \begin{equation}
% \begin{aligned}
% &f1\_reward = \begin{cases}
% -\frac{L_{predict} }{L},  real\_intervals \neq None \\
% f1(predict\_intervals,\ real\_intervals),  otherwise
% \end{cases}
% \\
% &format\_reward =
% \begin{cases}
% 1, & \langle think \rangle \langle /think \rangle\ \left[ \left[ start1, start2 \right] \right ....] \\
% 0, & otherwise \\
% \end{cases} \\
% &reward = 0.9 \times f1\_reward + 0.1 \times format\_reward 
% \end{aligned}
% \end{equation}
% }
% {
% \small
% \[
% \begin{aligned}
% &f1\_reward = \begin{cases}
% -\frac{L_{predict} }{L},  real\_intervals \neq None \\
% f1(predict\_intervals,\ real\_intervals), otherwise
% \end{cases}
% \\
% &format\_reward =
% \begin{cases}
% 1, & \langle think \rangle \langle /think \rangle\ \left[ \left[ start1, start2 \right] \right ....] \\
% 0, & otherwise \\
% \end{cases} \\
% &reward = 0.9 \times f1\_reward + 0.1 \times format\_reward 
% \end{aligned}
% \]
% }
{
\small
\[
\begin{aligned}
&f1\_reward = \begin{cases}
-0.5,  real=None \ and \  predict \neq None\\
0.5,  real=None \  and \ predict=None\\
f1(predict,\ real), otherwise
\end{cases}
\\
&format\_reward =
\begin{cases}
1, & \langle think \rangle \langle /think \rangle\ \left[ \left[ start1, start2 \right] \right ....] \\
0, & otherwise \\
\end{cases} \\
&reward = 0.9 \times f1\_reward + 0.1 \times format\_reward 
\end{aligned}
\]
}

\begin{figure}[tbp]
    \centering
    \includegraphics[width=0.45\textwidth]{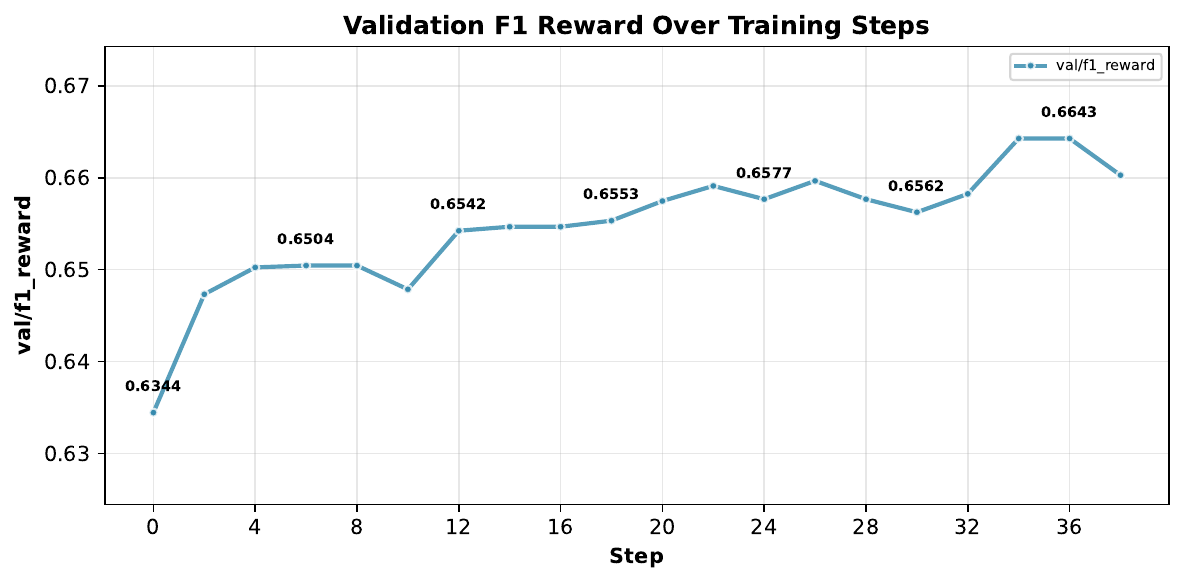} % 替换为你的文件名
    \caption{Reward during RL stage.}
    \label{fig:reward}
\end{figure}

We choose to incorporate an F1-based reward for its simplicity and effectiveness, while also including a format reward to guide the model in generating outputs that adhere to the required structure and enhance its reasoning ability. As described in the above formula, for the $f1\_reward$, we consider two aspects. For windows with anomalies, we directly calculate the F1 score as the reward. For windows without anomalies, the F1 score is 0 regardless of the model's response. Since we do not want the model to falsely identify anomalies in normal windows, we give a reward of 0.5 for correct predictions and -0.5 for incorrect predictions. We also demonstrate the effectiveness of this reward model through subsequent experiments. As shown in Figure \ref{fig:reward}, even a small number of training steps results in significant performance gains in TSAD, demonstrating the effectiveness of the final stage.

\subsection{Fixed Length Training and Adaptive Length Inference}
\label{sec:fixed}

% 传统的时序大模型构建方法，一般采用随机长度的时序训练模型，从而使得模型能够处理尽可能多样性的时序。然而，这种方式的弊端是训练数据量级巨大。然而不幸的是，即便再大量级的数据也无法涵盖所有长度的时序，某些场景下，需要对very long time series进行分析。因此，一个好的TS-VLM，应该能够对于任意长度的时序完成异常检测任务。

% 因此，paper提出了一种新的训练评估schema。如图\ref{fig:size}所示，我们采用fixed length的training 以及 dynamic length的evaluation。具体来说，训练阶段使用的time series length相同均为200，在evaluation阶段，对于不同特征的时序，我们可以动态的选择不同的长度，并通过放缩放缩到长度200，在计算F1 分数以及评估阶段，再放缩回原始长度。通过这种方式，不仅仅缩小了 训练数据量，同时使得模型可以适应任何长度、任何特征的time series。

Traditional methods for constructing time series large models generally use time series of random lengths for training, allowing the model to handle as much diversity as possible. However, the downside of this approach is that the training data size becomes enormous. Unfortunately, even with a massive amount of data, it is impossible to cover all lengths of time series, and some scenarios require the analysis of very long time series. Therefore, an effective TS-VLM should be able to perform anomaly detection on time series of any length.

\begin{figure}[tbp]
    \centering
    \includegraphics[width=0.48\textwidth]{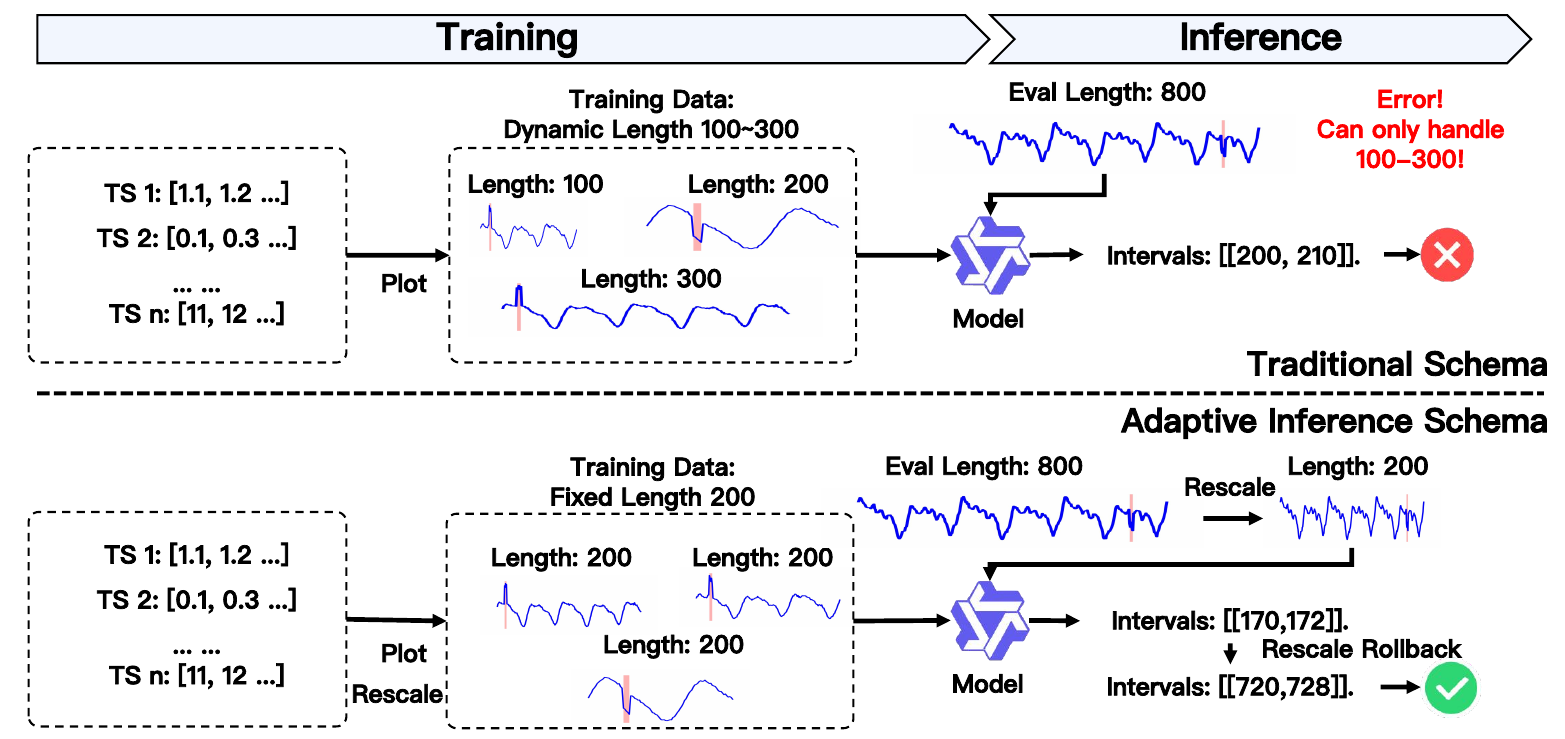} % 替换为你的文件名
    \caption{Fixed length training and adaptive length inference schema.}
    \label{fig:resize}
\end{figure}

To address this, as illustrated in Figure \ref{fig:resize}, we adopt a fixed length training and adaptive length inference schema. Specifically, in the training phase, all time series lengths in images are kept the same at 200 (after rescale). During the inference phase, for time series with different characteristics, we dynamically select different lengths and rescale time series images to a length of 200 (e.g., $[0, 1, \ldots, 800] \rightarrow [0, 0.25, \ldots, 200]$). After inference, we rescale them back to their original lengths (e.g., $[0, 0.25, \ldots, 200] \rightarrow [0, 1, \ldots, 800]$). This method not only reduces the amount of training data but also enables the model to adapt to time series of any length and with any characteristics.

\label{vits}

\section{Evaluation}

\begin{table*}[tbp]
\centering
\small
\caption{Overall performance of different models on synthetic dataset. The VLM ViTs used is Qwen2.5-VL-7B. All Qwen series models we employ are instruct models. (* indicates textual time series input. All models of Qwen series are instruct models.)}
\label{tab:performance}
\resizebox{\textwidth}{!}{
\begin{tabular}{ccccccccccccccccccc} 
\toprule
 & \multicolumn{3}{c}{\textbf{Spike}} & \multicolumn{3}{c}{\textbf{Trend}} & \multicolumn{3}{c}{\textbf{Level}} & \multicolumn{3}{c}{\textbf{Frequency}} & \multicolumn{3}{c}{\textbf{Mixed}} & \multicolumn{3}{c}{\textbf{Overall}} \\
\cmidrule{2-19}
 & P & R & F1 & P & R & F1 & P & R & F1 & P & R & F1 & P & R & F1 & P & R & F1 \\
\midrule
\textbf{Qwen2.5-VL-7B*} 
 & 0.4496 & 0.0099 & 0.0194 & 0.3327 & 0.0316 & 0.0577 & 0.1660 & 0.0086 & 0.0164 & 0.0000 & 0.0000 & 0.0000 & 0.0000 & 0.0000 & 0.0000 & 0.1411 & 0.0117 & 0.0217 \\
\textbf{Qwen2.5-VL-7B}  
 & 0.5631 & 0.5490 & 0.5560 & 0.7019 & 0.0913 & 0.1616 & 0.5733 & 0.5819 & 0.5776 & 0.8993 & 0.3035 & 0.4538 & 0.6910 & 0.3107 & 0.4287 & 0.3813 & 0.4216 & 0.4004 \\
\textbf{Qwen2.5-VL-32B} 
 & 0.5436 & 0.6521 & 0.5929 & 0.6045 & 0.2652 & 0.3686 & 0.5398 & 0.6552 & 0.5919 & 0.6539 & 0.4432 & 0.5283 & 0.6825 & 0.2281 & 0.3419 & 0.3262 & 0.5216 & 0.4014 \\
\textbf{Qwen2.5-VL-72B} 
 & 0.6385 & 0.8296 & 0.7216 & 0.6474 & 0.6431 & \underline{0.6453} & 0.5841 & 0.8556 & 0.6943 & 0.7757 & 0.7513 & 0.7633 & 0.7144 & 0.4885 & 0.5802 & 0.4071 & 0.7671 & 0.5319 \\
\textbf{InternVL3-8B}   
 & 0.5273 & 0.6147 & 0.5677 & 0.6634 & 0.5098 & 0.5766 & 0.4416 & 0.5713 & 0.4981 & 0.7318 & 0.5192 & 0.6075 & 0.8024 & 0.5109 & 0.6243 & 0.3795 & 0.5661 & 0.4544 \\
\textbf{InternVL3-14B}  
 & 0.6312 & 0.6813 & 0.6553 & 0.6759 & 0.5650 & 0.6155 & 0.5680 & 0.6342 & 0.5993 & 0.7861 & 0.7081 & 0.7450 & 0.8059 & 0.5109 & \underline{0.6254} & 0.3785 & 0.6406 & 0.4758 \\
\textbf{InternVL3-38B}  
 & 0.5806 & 0.7289 & 0.6464 & 0.6053 & 0.6012 & 0.6032 & 0.4907 & 0.7799 & 0.6024 & 0.7543 & 0.7847 & \underline{0.7692} & 0.7509 & 0.4666 & 0.5756 & 0.3035 & 0.7078 & 0.4248 \\
\textbf{GPT-4o-mini}    
 & 0.4326 & 0.4628 & 0.4472 & 0.6194 & 0.4614 & 0.5289 & 0.3883 & 0.4420 & 0.4134 & 0.7001 & 0.5845 & 0.6371 & 0.6801 & 0.4885 & 0.5686 & 0.2173 & 0.4805 & 0.2993 \\
\textbf{GPT-4o}         
 & 0.6938 & 0.7533 & \underline{0.7223} & 0.8064 & 0.3674 & 0.5048 & 0.7024 & 0.7415 & \underline{0.7214} & 0.8836 & 0.5438 & 0.6733 & 0.7634 & 0.4447 & 0.5620 & 0.5710 & 0.6284 & \underline{0.5983} \\
\textbf{ViTs}           
 & 0.9565 & 0.8835 & \textbf{0.9185} 
 & 0.9010 & 0.6949 & \textbf{0.7846} 
 & 0.9687 & 0.9404 & \textbf{0.9543} 
 & 0.9526 & 0.8336 & \textbf{0.8891} 
 & 0.9516 & 0.6668 & \textbf{0.7842} 
 & 0.8791 & 0.8380 & \textbf{0.8581} \\
\bottomrule
\end{tabular}
}
\end{table*}

\begin{table*}[t]
\centering
\footnotesize
\caption{Zero-shot performance of different methods on public datasets. Zero-shot setting is explained in detail in Section \ref{sec:expsetting}.}
\label{tab:exp2}
\resizebox{\textwidth}{!}{
\begin{tabular}{cllllllllllll} 
\toprule
\multicolumn{1}{c}{\multirow{2}{*}{\textbf{Methods}}} & \multicolumn{3}{c}{\textbf{KPI}}                                                                  & \multicolumn{3}{c}{\textbf{Yahoo}}                                                                & \multicolumn{3}{c}{\textbf{WSD}}                                                                  & \multicolumn{3}{c}{\textbf{Average}}                                                               \\ 
\cmidrule{2-13}
\multicolumn{1}{c}{}                                  & \multicolumn{1}{c}{\textbf{P}} & \multicolumn{1}{c}{\textbf{R}} & \multicolumn{1}{c}{\textbf{F1}} & \multicolumn{1}{c}{\textbf{P}} & \multicolumn{1}{c}{\textbf{R}} & \multicolumn{1}{c}{\textbf{F1}} & \multicolumn{1}{c}{\textbf{P}} & \multicolumn{1}{c}{\textbf{R}} & \multicolumn{1}{c}{\textbf{F1}} & \multicolumn{1}{c}{\textbf{P}} & \multicolumn{1}{c}{\textbf{R}} & \multicolumn{1}{c}{\textbf{F1}}  \\ 
\midrule
\textbf{Spot \cite{spot}}                                                  & 0.9661                         & 0.2213                         & 0.3603                          & 0.5723                         & 0.3281                         & 0.4174                          & 0.9475                         & 0.3156                         & 0.4728                          & 0.8280                         & 0.2883                         & 0.4168                           \\
\textbf{SubLOF \cite{breunig2000lof}}                                                & 0.7191                         & 0.7488                         & 0.7024                          & 0.5213                         & 0.8361                         & 0.5678                          & 0.6523                         & 0.8806                         & 0.6993                          & 0.6309                         & 0.8218                         & 0.6565                           \\
\textbf{Anomaly-Transformer \cite{anomaly-transfomer}}                                   & 0.6917                         & 0.7135                         & 0.6701                          & 0.5450                         & 0.8022                         & 0.5717                          & 0.7157                         & 0.8329                         & 0.6473                          & 0.6508                         & 0.7828                         & 0.6297                           \\
\textbf{DCDetector \cite{yang2023dcdetector}}                                            & 0.5495                         & 0.7114                         & 0.5862                          & 0.3709                         & 0.8219                         & 0.4597                          & 0.3096                         & 0.8349                         & 0.3830                          & 0.4100                         & 0.7894                         & 0.4763                           \\
\textbf{TranAD \cite{tuli2022tranad}}                                                & 0.8667                         & 0.7768                         & \textbf{0.7855}                  & 0.6770                         & 0.8326                         & 0.6698                          & 0.7549                         & 0.8583                         & 0.6988                          & 0.7662                         & 0.8225                         & 0.7180                           \\
\textbf{TimesNet \cite{wu2022timesnet}}                                              & 0.7245                         & 0.7988                         & 0.7183                          & 0.6922                         & 0.8491                         & \underline{0.7101}                  & 0.8345                         & 0.9598                         & \textbf{0.8617}                 & 0.7504                         & 0.8692                         & \underline{0.7633}                   \\
\textbf{ViTs}                                       & \multicolumn{1}{c}{0.7656}     & \multicolumn{1}{c}{0.7590}     & \underline{0.7626}                 & 0.7444                         & 0.9558                         & \textbf{0.8373}                 & 0.7645                         & 0.9565                         & \underline{0.8497}                  & 0.7581                         & 0.8904                         & \textbf{0.8142}                  \\
\bottomrule
\end{tabular}
}
\end{table*}

% \subsection{Experiment Settings}

% Please refer to Section \ref{sec:expsetting}.

\subsection{Datasets and Baselines}

% For the evaluation dataset, we utilize 2k artificially generated time series, where anomalies are inserted with a probability of 0.7 in each sequence. In terms of evaluation metrics, we employ the commonly used TSAD benchmarks: precision, recall, and point-adjusted F1 score. For baseline comparisons, among open-source models, we select the current SOTA VLMs, including the Qwen2.5-VL \cite{qwen2.5-VL} series and InternVL \cite{chen2024internvl} series, while for proprietary models, we choose the leading SOTA models GPT-4o-mini and GPT-4o \cite{achiam2023gpt}.

% For the evaluation dataset, we utilize 2k synthetic time series, where anomalies are randomly inserted into the time series. In terms of evaluation metrics, we employ commonly used TSAD benchmarks: precision, recall, and point-adjusted F1 score. For baseline comparisons, among open-source models, we select the current SOTA VLMs, including the Qwen2.5-VL \cite{qwen2.5-VL} series and InternVL \cite{chen2024internvl} series, while for proprietary models, we choose the leading SOTA models GPT-4o-mini and GPT-4o \cite{achiam2023gpt}. Additionally, we compare ViTs with SOTA methods on public datasets (KPI, Yahoo).

We evaluated the performance of ViTs using both synthetic data and public datasets. The synthetic dataset contains 2k samples, with anomalies randomly inserted into the time series. The public datasets used include the commonly used benchmarks KPI, Yahoo, and WSD. In terms of metrics, we employ commonly used TSAD indicators: precision, recall, and point-adjusted F1 score. For baseline comparisons, among open-source models, we select the current SOTA VLMs, including the Qwen2.5-VL \cite{qwen2.5-VL} series and InternVL \cite{chen2024internvl} series, while for proprietary models, we choose the leading SOTA models GPT-4o-mini and GPT-4o \cite{achiam2023gpt}. Additionally, we compare ViTs with SOTA TSAD methods. Detailed experimental settings can be found in Section \ref{sec:expsetting}.

% 真实评测中LLM-based 的方法很难输出异常概率或者异常分数，而人们比较关注异常的确信度。因此为了解决这个问题，我们提出了滑动窗口为基础的异常分数。具体来说，通过滑动窗口，对于同一个点，利用LLM判断多次，LLM判断为异常的次数最终作为异常分数参与到最终F1的计算中。
In real-world evaluations, LLM-based methods often struggle to output anomaly probabilities or scores, while users are more concerned with the confidence level of the anomalies. To address this issue, we propose a sliding window-based anomaly score. Specifically, by using a sliding window, each point in the time series is evaluated multiple times by the LLM. The number of times the LLM identifies a point as an anomaly is used as the anomaly score, which is then incorporated into the final F1 score calculation.

\subsection{Overall Performance}

% 人造数据集的评估结果如Table \ref{tab:performance}所示。可以看出，不论是开源模型还是闭源模型，随着模型size的增大，模型在各类异常上的检测效果都得到明显提升。甚至Qwen2.5-VL-72B的能力已经和GPT-4o相近，某些异常上的检测效果甚至超过GPT-4o。然而，即便是当前的SOTA模型，F1分数仍然不足0.6。而ViTs仅使用7B的参数，经过总计15k条数据，便可以将平均F1分数提升25\%以上，并且在各类异常上表现均衡。

% 公开数据集上的评估结果如图\ref{tab:exp2}所示。可以看出，尽管ViTs 纯在合成数据上进行训练，但是ViTs却能够zero-shot的检测公开数据集中的异常，并且效果超过了当前SOTA，平均F1分数达到了0.8左右。这证明了ViTs 数据构造的多样性以及训练策略的有效性，同时证明了ViTS有很好的的泛化能力。

The evaluation results on the synthetic dataset are shown in Table \ref{tab:performance}. It can be observed that for both open-source and proprietary models, the TSAD performance for various types of anomalies improves significantly with an increase in model size. Notably, the capabilities of Qwen2.5-VL-72B are comparable to those of GPT-4o, with superior performance for certain anomalies. However, even the current SOTA models have an F1 score that falls short of 0.6. In contrast, ViTs, with only 7B parameters and trained with 15k data samples, achieves an average F1 score improvement of over 25\% and exhibits balanced performance across various types of anomalies.

The evaluation results on public datasets are shown in Table \ref{tab:exp2}. Despite being trained solely on synthetic data, ViTs is able to zero-shot detect anomalies in public datasets, outperforming the current SOTA and achieving an average F1 score of approximately 0.8. This demonstrates the diversity of ViTs' data construction and the effectiveness of its training strategy, while also proving ViTs' excellent generalization capabilities.

\begin{table}[tbp]
\centering
\footnotesize
\caption{Overall performance of different settings on synthetic data. Full results are shown in Table \ref{tab:fullablation}.}
\label{tab:ablation}
\resizebox{0.5\textwidth}{!}{
\begin{tabular}{ccccccc} 
\toprule
\textbf{Setting}                & Variant           & P      & R      & F1      \\ 
\midrule
\textbf{Naive}                  &                   & 0.3813 & 0.4216 & 0.4004 \\
\textbf{\#1 SFT-\#2 SFT}        & \#1 Frozen LLM    & 0.9388 & 0.6280 & 0.7526 \\
\textbf{\#1 SFT-\#2 SFT}        & \#1 Full          & 0.9483 & 0.6602 & 0.7785 \\
\textbf{\#2 SFT}                &                   & 0.9345 & 0.6400 & 0.7597 \\
\textbf{\#3 RL}                 &                   &    0.5892    &   0.7101     &  0.6440      \\
\textbf{\#1 SFT-\#2 SFT-\#3 RL} & \#3 Full          & 0.8791 & 0.8380 & \textbf{0.8581} \\
\textbf{\#1 SFT-\#2 SFT-\#3 RL} & \#3 Frozen Vision & 0.8829 & 0.8224 & \underline{0.8516} \\
\bottomrule
\end{tabular}
}
\end{table}

\subsection{Ablation Study}

To demonstrate the effectiveness of Chain-of-TS fine-tuning strategy, we compared different strategies. The results are shown in Table \ref{tab:ablation}. It can be observed that both the initial SFT steps and the third RL step contribute to improvements in the final TSAD performance, with the third stage RL providing nearly a 10\% boost. Removing any of the stages significantly degrades performance. Overall, SFT injects new knowledge into the model, while RL helps the model explore and utilize this knowledge for better results. Additionally, we evaluated the impact of freezing LLM parameters during the SFT stage. It can be seen that freezing LLM parameters in the first stage negatively affects the results, as it creates a modality gap. Freezing the vision parameters during RL has little impact since the reasoning and exploration capabilities are primarily provided by the LLM backbone. Therefore, in the third stage, we chose to freeze the vision parameters to accelerate training efficiency.

% \begin{table}[tbp]
% \centering
% \caption{Overall performance of different settings on synthetic data. Full results are shown in Table \ref{tab:fullablation}.}
% \label{tab:ablation}
% \resizebox{0.5\textwidth}{!}{
% \begin{tabular}{ccccccc} 
% \toprule
% \textbf{Setting}                & Variant           & P      & R      & F1      \\ 
% \midrule
% \textbf{Naive}                  &                   & 0.3813 & 0.4216 & 0.4004 \\
% \textbf{\#1 SFT-\#2 SFT}        & \#1 Frozen LLM    & 0.9388 & 0.6280 & 0.7526 \\
% \textbf{\#1 SFT-\#2 SFT}        & \#1 Full          & 0.9483 & 0.6602 & 0.7785 \\
% \textbf{\#2 SFT}                &                   & 0.9345 & 0.6400 & 0.7597 \\
% \textbf{\#3 RL}                 &                   &        &        &        \\
% \textbf{\#1 SFT-\#2 SFT-\#3 RL} & \#3 Full          & 0.8791 & 0.8380 & \textbf{0.8581} \\
% \textbf{\#1 SFT-\#2 SFT-\#3 RL} & \#3 Frozen Vision & 0.8829 & 0.8224 & \underline{0.8516} \\
% \bottomrule
% \end{tabular}
% }
% \end{table}

\subsection{Fixed Length Training and Adaptive Length Inference}

% 在Section \ref{}中我们提出了fixed length training adaptive length evaluation的schema。为了验证该schema的有效性。我们使用数量相同的（10k）的训练数据（与stage 2阶段类型相同的TSAD QA）训练，分别使用fixed length 和dynamic length 训练，在相同的人造数据集上（length=200）评估结果如图\ref{}所示。可以看出，在对大多数类型的异常上，fixed length trainin的效果均好于dynamic length 。这很好理解，fixed length的training 可以使得VLM学习更多该length的样本，从而有更强的TSAD能力。

\begin{figure}[tbp]
    \centering
    \includegraphics[width=0.45\textwidth]{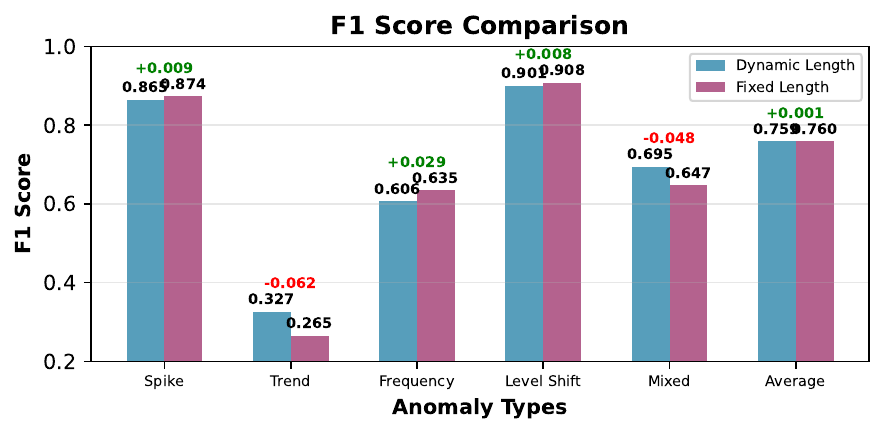} % 替换为你的文件名
    \caption{Comparison of fixed length training and dynamic length training.}
    \label{fig:fixed}
\end{figure}

% 除了效果提升之外，fixed length training adaptive length evaluation可以让模型具备处理任意长度时序的能力，更适配当前的真实场景。
In Section \ref{sec:fixed}, we proposed a new fiexed length training and adaptive inference schema with time series image rescaling. To verify the effectiveness of fixed length training, we used the same amount of training data (10k TSAD QA of the same type as in stage 2) and trained models using both fixed length and dynamic length training approaches. The evaluation results on the same synthetic dataset (length = 200) are shown in Figure \ref{fig:fixed}. It can be seen that for most types of anomalies, fixed length training outperforms dynamic length training. This is understandable, as fixed length training allows the VLM to learn more samples of that length, thereby enhancing its TSAD capabilities.

%为了验证 adpative length inference的有效性，我们对公开数据集使用fixed length training 之后的模型进行评测，结果如表\ref{}所示。可以看出，对于不同的dataset，适合的最佳length不一样。因此，有了adaptive length inference，我们可以针对不同的场景选择不同的length进行评测，并且不需要重新训练模型。

\begin{table}[t]
\centering
\footnotesize
\caption{Performance on KPI and Yahoo datasets with different window sizes.}
\label{tab:aiops_yahoo_window}
\resizebox{0.5\textwidth}{!}{
\begin{tabular}{cccccc}
\toprule
\textbf{Dataset} & \textbf{Window} & \textbf{F1} & \textbf{P} & \textbf{R} & \textbf{VUS\_ROC} \\
\midrule
KPI  & 100 & 0.5630 & 0.4090 & 0.9018 & 0.6166 \\
KPI  & 200 & 0.7626 & 0.7656 & 0.7590 & 0.6106 \\
Yahoo  & 50  & 0.8373 & 0.7444 & 0.9558 & 0.9389 \\
Yahoo  & 100 & 0.8324 & 0.7378 & 0.9552 & 0.9304 \\
\bottomrule
\end{tabular}
}
\end{table}

% To validate the effectiveness of adaptive length inference with time series image rescaling, we evaluated models trained with fixed length on public datasets. The results are shown in Table \ref{tab:aiops_yahoo_window}. It can be seen that the optimal length varies for different datasets. Therefore, with adaptive length inference, we can choose different lengths for evaluation based on different scenarios without the need to retrain the model.
To verify the effectiveness of adaptive length inference enabled by time series image rescaling, we evaluated models trained with a fixed length on several public datasets. The corresponding results are presented in Table~\ref{tab:aiops_yahoo_window}. As shown, the optimal sequence length varies across different datasets. Therefore, with adaptive length inference, models can flexibly select suitable lengths for evaluation in different scenarios without requiring retraining.

% In addition to performance improvement, fixed-length training with adaptive-length evaluation enables models to handle time series of any length, making it more suitable for real-world scenarios.

\begin{figure}[tbp]
  \centering
  % 第1行
  \begin{subfigure}[b]{0.22\textwidth}
    \includegraphics[width=\linewidth]{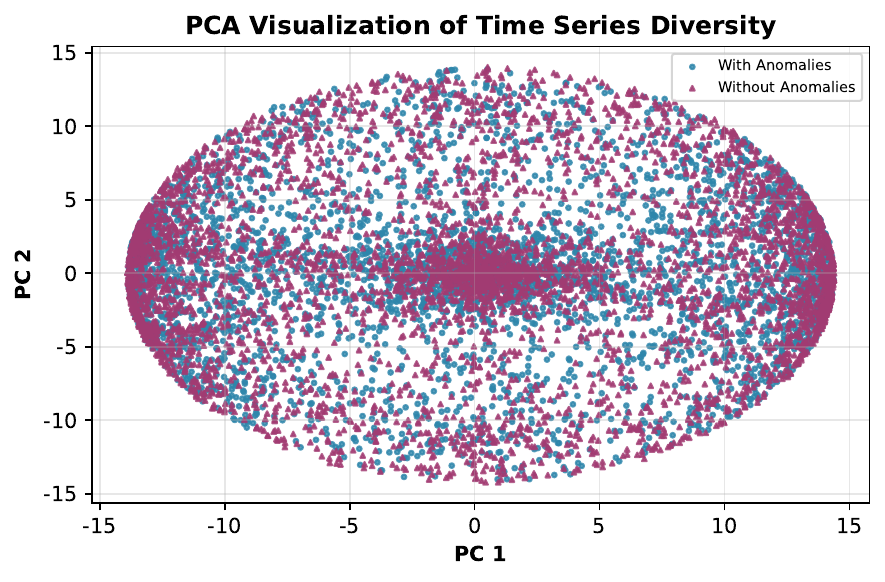}
    \caption{PCA}
  \end{subfigure}
  \hfill
  \begin{subfigure}[b]{0.22\textwidth}
    \includegraphics[width=\linewidth]{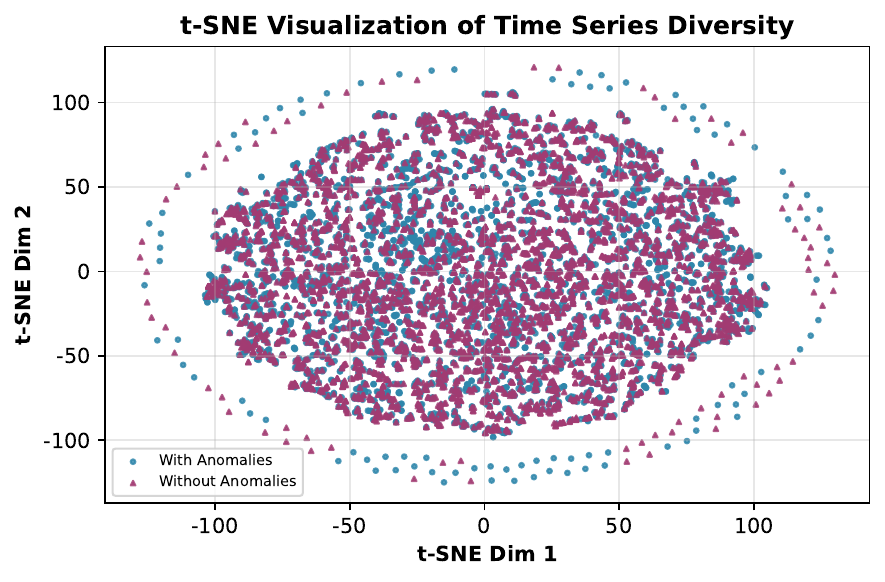}
    \caption{t-SNE}
  \end{subfigure}

  \vskip\baselineskip  % 两行之间加一点垂直间距

  % 第2行
  \begin{subfigure}[b]{0.22\textwidth}
    \includegraphics[width=\linewidth]{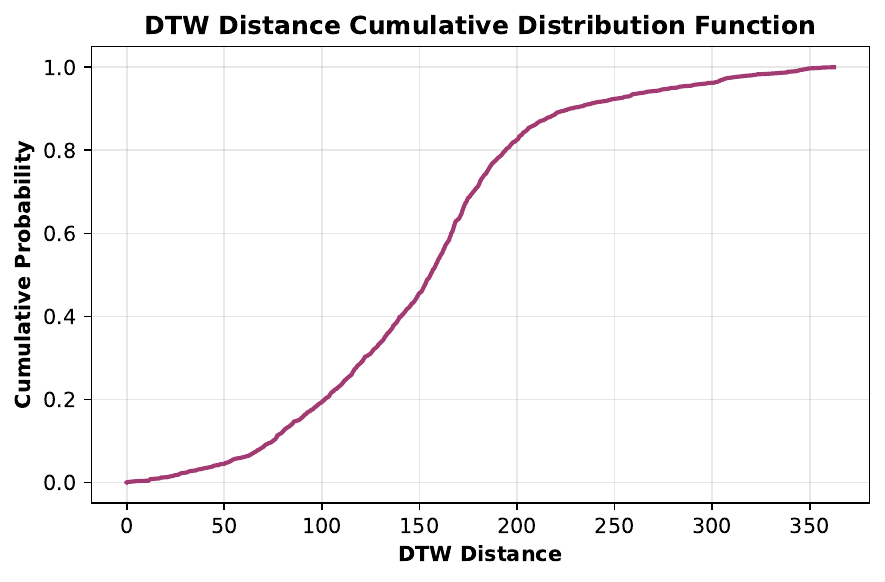}
    \caption{CDF}
  \end{subfigure}
  \hfill
  \begin{subfigure}[b]{0.22\textwidth}
    \includegraphics[width=\linewidth]{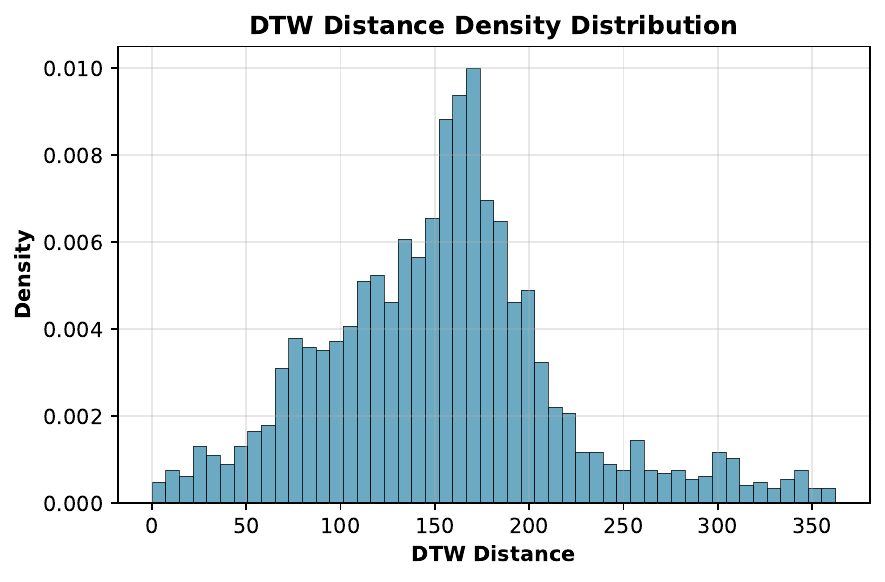}
    \caption{Distribution}
  \end{subfigure}

  \caption{Visualization of data diversity.}
  \label{fig:datadiversity}
\end{figure}

\subsection{Data Diversity}

% 在LLM领域，训练数据是基石，其知识丰富度和多样性决定了模型能完成的任务类型以及完成任务的准确率。在TSAD领域同样如此，只有模型见过足够丰富的time seires 数据才能保证模型在看到新的不同的time series 时也能够有效handle。尽管我们在上文中已经证明了通过STL合成的数据，其中seasoanl 部分可以逼近任意连续函数，并且误差有界。这里，为了进一步证明人造数据的多样性，我们分别对于10k条样本进行了 t-SNE和PCA处理，并可视化。结果如图\ref{}所示。可以看出正常时序和异常时序都均匀的分布在各个地方，并且形成一个圆形轮廓。这是因为所有时序都经过了z-score标准化，因此是有界限的。
% 此外，我们还随机采样了2000个pair 计算其dtw距离，并作出CDF和分布图，可以看出，大多数时序之间的距离相对较远，这进一步证明了时序的多样性。

In the field of LLMs, training data serves as the cornerstone, with its richness and diversity determining the types of tasks a model can perform and the accuracy with which it can complete them. The same holds true for TSAD; only if a model has encountered sufficiently rich time series data can it effectively handle new and different time series. Although we have previously demonstrated that data synthesized through STL, where the seasonal component can approximate any continuous function with bounded error, further evidence of the diversity of synthetic data is provided here.

For this purpose, we perform t-SNE and PCA processing on 10k samples and visualize the results. As shown in Figure \ref{fig:datadiversity}, normal and anomalous time series are evenly distributed, forming a circular outline. This is expected because all time series were standardized using z-score normalization, ensuring they are bounded.

Additionally, we randomly sampled 2000 pairs and calculated their DTW distances, plotted as a CDF and distribution graph. The results indicate that most time series are relatively distant from each other, further proving the diversity of the time series.

\subsection{Reward Model}

\begin{figure}[tbp]
    \centering
    \includegraphics[width=0.45\textwidth]{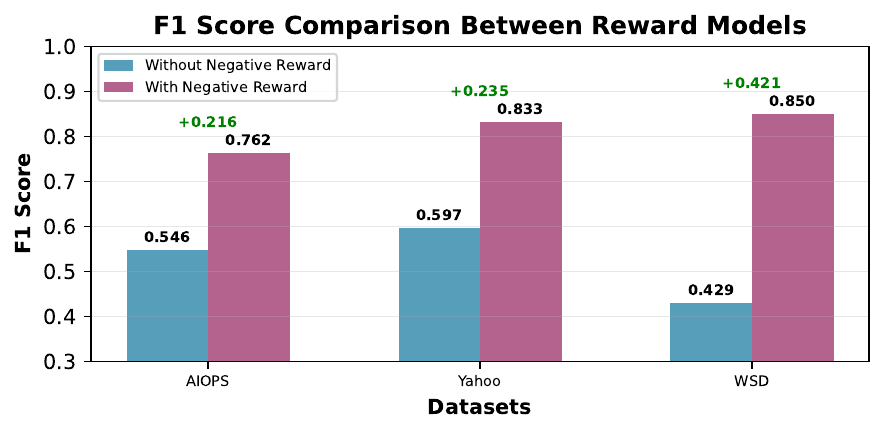} % 替换为你的文件名
    \caption{Comparison of different reward model.}
    \label{reward}
\end{figure}

% RL作为Chain-of-TS 策略中的重要一环，扮演了重要角色。然而RL中很重要的一个环节是reward model的设计。DeepSeek R1证明了简单的基于规则的reward model便可以达到相对较好的效果。因此我们设计了F1分数为基础的reward。然而，我们不仅仅使用F1分数，同时还真对无异常的区间设计负数的reward。这是因为，如果不加以限制，对于无异常窗口，任何预测结果F1分数均为0. 这会导致，模型RL过程倾向于不断采样区间满足异常窗口的F1值相对较大，导致对于无异常的窗口也会大概率预测异常。这里，我们比较了负数reward添加对于结果的影响，如图\ref{reward}所示。可以看出，不加负数reward会增加FN的个数，进而导致precision和F1分数很低。

RL plays an important role as a key part of the Chain-of-TS strategy. However, a critical element in RL is the design of the reward model. DeepSeek R1 \cite{guo2025deepseek} has demonstrated that a simple rule-based reward model can achieve relatively good results. Therefore, we designed a reward model based on the F1 score. However, we did not solely use the F1 score; we also designed a negative reward for segments without anomalies. This is because, without restrictions, any prediction result for an anomaly-free window would yield an F1 score of 0. This would cause the model to bias towards sampling segments that satisfy the anomaly window's F1 value, leading to a high likelihood of predicting anomalies in windows without anomalies. 

Here, we compared the impact of adding a negative reward on the results, as shown in Figure \ref{reward}. It can be seen that not adding a negative reward increases the number of false positives, thereby significantly reducing precision and the F1 score.

% \begin{figure}[tbp]
%   \centering
%   % 第1行
%   \begin{subfigure}[b]{0.22\textwidth}
%     \includegraphics[width=\linewidth]{pic/pca.pdf}
%     \caption{PCA}
%   \end{subfigure}
%   \hfill
%   \begin{subfigure}[b]{0.22\textwidth}
%     \includegraphics[width=\linewidth]{pic/t-sne.pdf}
%     \caption{t-SNE}
%   \end{subfigure}
%   \hfill
%   \begin{subfigure}[b]{0.22\textwidth}
%     \includegraphics[width=\linewidth]{pic/dtw_distance_cdf.pdf}
%     \caption{CDF}
%   \end{subfigure}
%   \hfill
%   \begin{subfigure}[b]{0.22\textwidth}
%     \includegraphics[width=\linewidth]{pic/dtw_distance_density.pdf}
%     \caption{Distribution}
%   \end{subfigure}

%   \caption{Visualization of data diversity.}
%   \label{fig:datadiversity}
% \end{figure}

\label{evaluation}

\section{Conclusion}

This work presents \textbf{ViTs}, a novel VLM that significantly advances the SOTA in TSAD. ViTs introduces an STL-based periodic data generation method and the Chain-of-TS fine-tuning strategy. Additionally, we propose a new fixed length training and adaptive length inference schema. Extensive experiments across multiple datasets demonstrate that ViTs outperforms SOTA by substantial margins. 

% The success of our framework not only addresses current limitations in VLM-based TSAD but also opens new research directions for multimodal time series analysis.

\label{conclusion}

\bibliographystyle{acm} % 参考文献样式
\bibliography{sample-base}

\appendix

\section{Related Work}

% time-llm aaai25chattime  llmad  chatts
% LLM由于其强大的文本理解能力和推理能力因此被大量应用在time series上。（1）LLM fortime series forcasting。其中有些方法利用LLM的语义知识，通过prompt engineering的方式，利用few-shot等trick使得LLM推理出预测数据。包括Time-llm ChatTime在内的一些其他方法，尝试利用大语言模型的参数知识，通过微调增强LLM Backbone forecast的能力。（2）LLM for TSAD：LLMAD，AnomalyLLM，sigllm等方法都在尝试利用prompt engineering，告诉LLM什么样的数据是什么类型的异常，目前并没有针对TSAD进行微调的方法。（3） LLM for Time Series Reasoning：ChatTS通过额外的Time series encoder以及大量的数据提升LLM时序分类等各项通用任务的能力，。

\textbf{LLMs for Time Series}\ \  LLMs have gained significant traction in time series analysis due to their exceptional textual reasoning capabilities, with applications spanning three main directions: (1) Time Series Forecasting: Some methods \cite{llmforcastprompt} leverage LLMs' semantic knowledge through prompt engineering techniques (e.g., few-shot learning), while approaches like Time-LLM \cite{timellm} and ChatTime \cite{wang2025chattime} enhance forecasting performance by fine-tuning the LLM backbone to better utilize its parametric knowledge \cite{llmforcastfintuning}; (2) Time Series Anomaly Detection: Methods such as LLMAD \cite{llmad}, AnomalyLLM \cite{anomalyllm}, and SigLLM \cite{sigllm} primarily employ prompt engineering to describe anomaly patterns, though notably no current approach has investigated dedicated TSAD-oriented fine-tuning of LLMs; (3) Time Series Reasoning: Methods like ChatTS \cite{xie2024chatts} demonstrate improved performance on time series classification and related tasks by incorporating specialized time series encoders and large-scale temporal data training, thereby extending LLMs' capabilities beyond their original text-based design.

% 最近越来越多的方法尝试使用VLM分析time series。AnoLLM通过大量的实验给出如何使用prompt engineering达到更好的TSAD效果。MLLM评估了当前使用multimodal LLM 进行TSAD的效果。总的来说，VLM进行TSAD有得天独厚的优势，更符合人类的思维方式，有更好的发展前景。

\textbf{VLMs for Time Series} \ \ Recent research has witnessed a growing trend of employing VLMs for time series analysis. Notable works include AnoLLM \cite{anollm}, which systematically investigates how prompt engineering strategies can optimize TSAD performance through extensive experimentation, and MLLM \cite{mllm}, which comprehensively evaluates the current capabilities of multimodal LLMs in TSAD tasks. Compared to conventional approaches, VLMs demonstrate unique advantages for TSAD applications - their multimodal reasoning paradigm better aligns with human cognitive processes, offering superior interpretability and greater potential for future development in this domain.
\label{relatedwork}

\section{Anomaly Types}
\label{sec:ano}

(1) \textbf{Spike Anomaly:} The most common type of anomaly, characterized by a sudden sharp increase or decrease in the time series value, followed by a return to the normal range after a short duration. (2) \textbf{Trend Anomaly:} Indicates a change in the overall trend of the time series, which may manifest as an alteration in the rate of increase/decrease or a reversal in trend direction (e.g., shifting from an upward to a downward trend). (3) \textbf{Level Anomaly:} Refers to an abrupt shift in the baseline level of the time series, after which the values fluctuate around this new baseline. (4) \textbf{Frequency Anomaly:} Represents a local disruption in the periodic pattern of the time series, typically involving significant changes in either the frequency or amplitude of fluctuations within a specific time window.

\section{Visualization of Different Plot Types}
\begin{figure*}[tbp]
  \centering

  % 第1行
  \begin{subfigure}[b]{0.22\textwidth}
    \includegraphics[width=\linewidth]{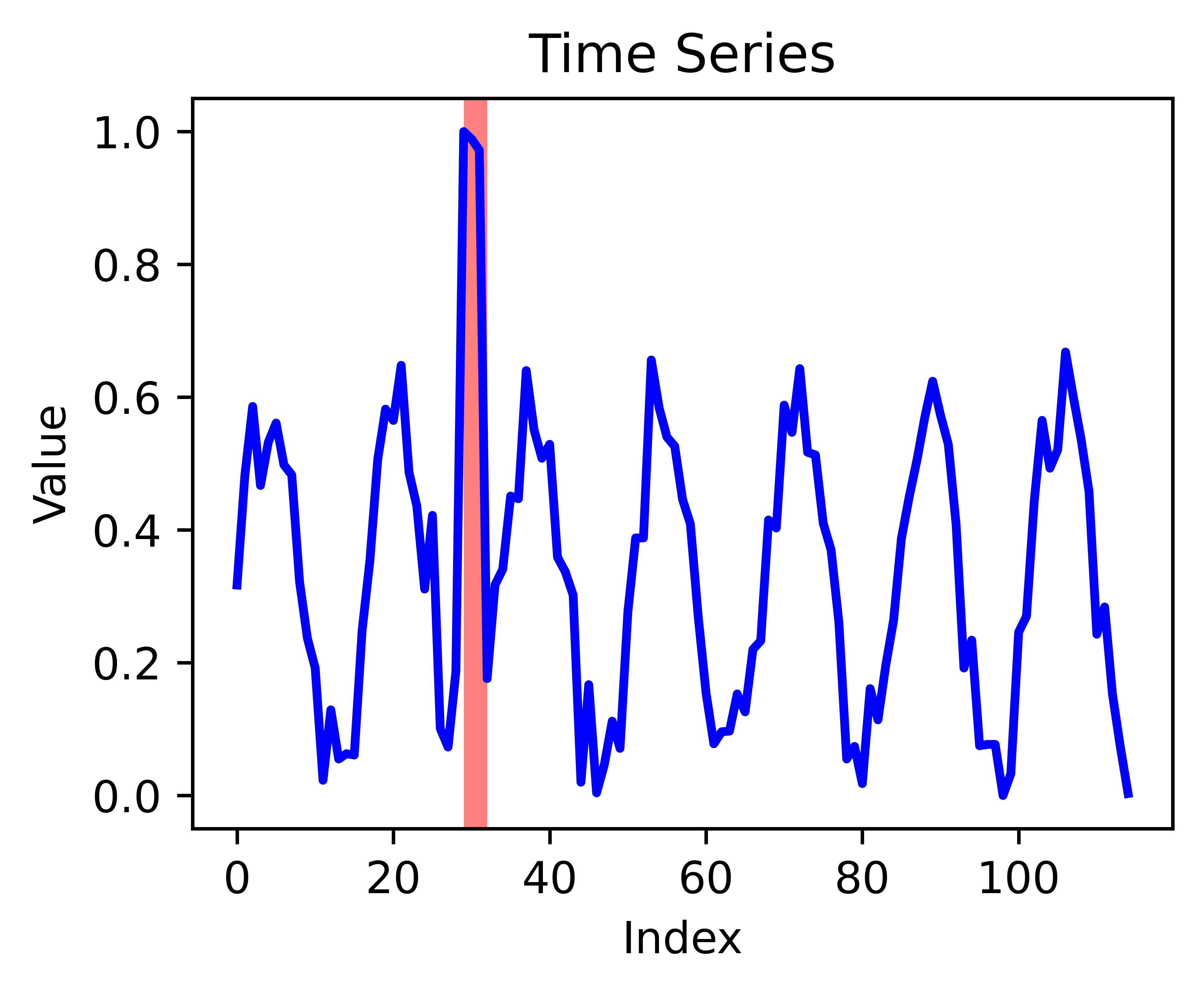}
    \caption{Line-Spike}
  \end{subfigure}
  \hfill
  \begin{subfigure}[b]{0.22\textwidth}
    \includegraphics[width=\linewidth]{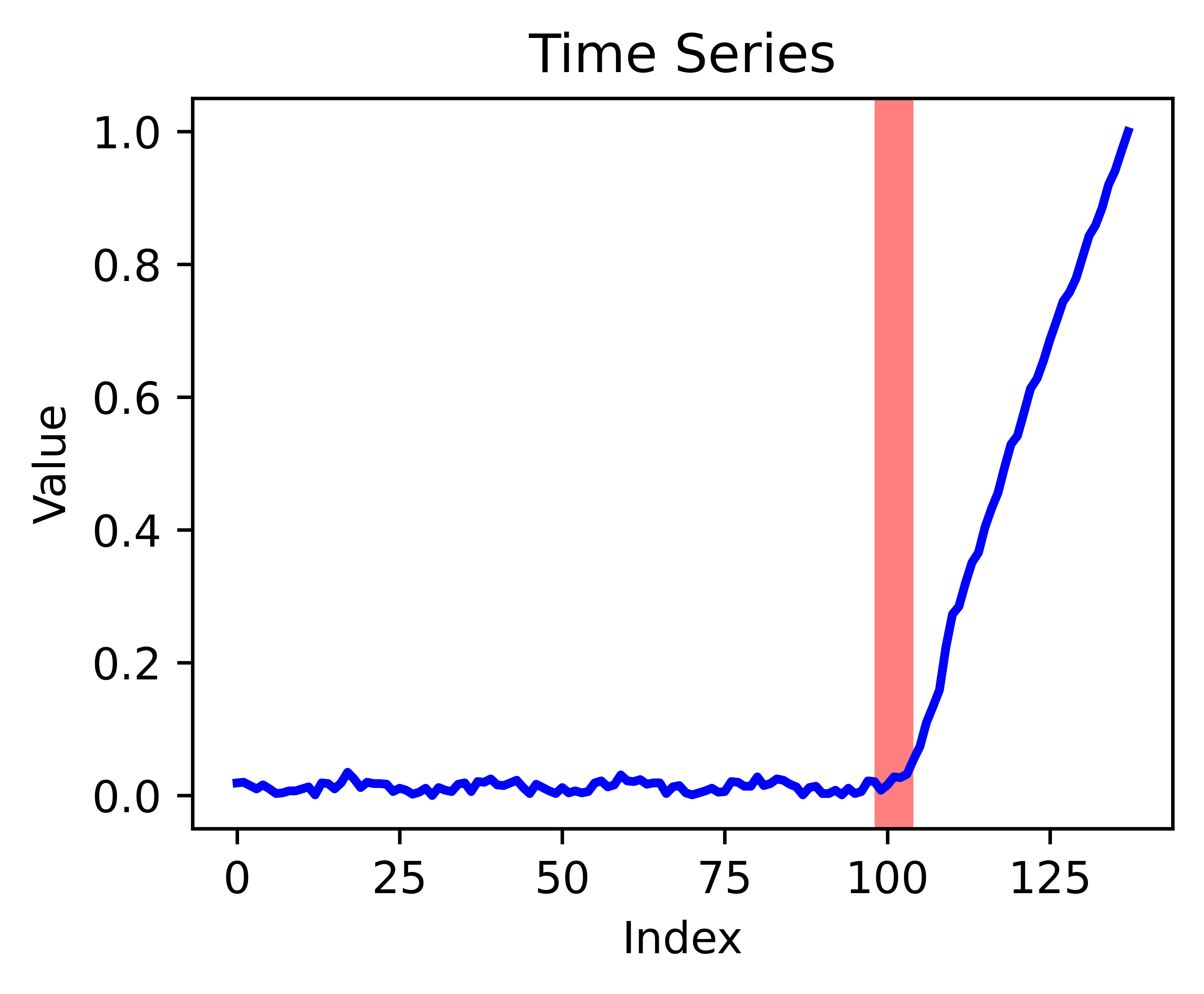}
    \caption{Line-Trend}
  \end{subfigure}
  \hfill
  \begin{subfigure}[b]{0.22\textwidth}
    \includegraphics[width=\linewidth]{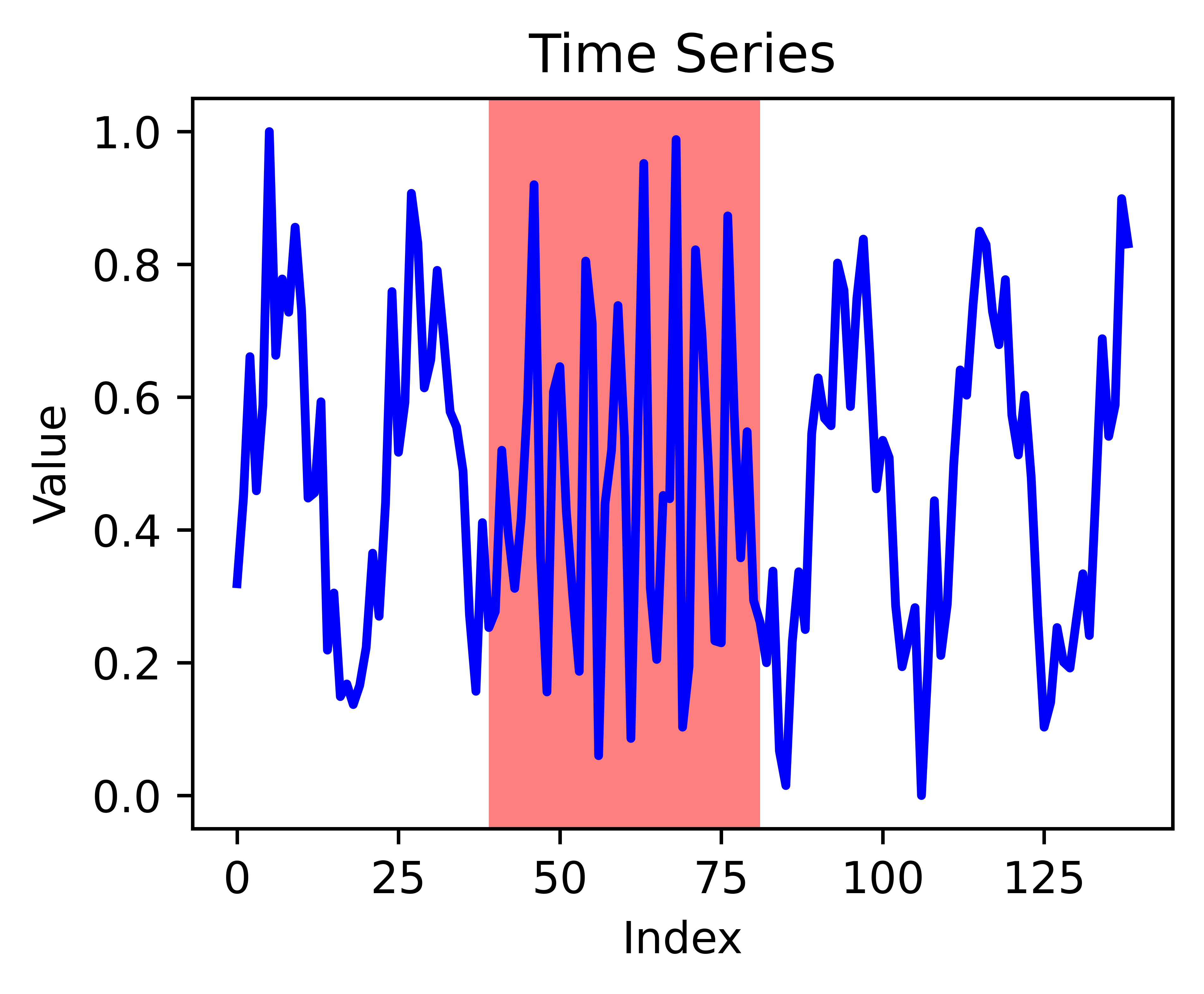}
    \caption{Line-Frequency}
  \end{subfigure}
  \hfill
  \begin{subfigure}[b]{0.22\textwidth}
    \includegraphics[width=\linewidth]{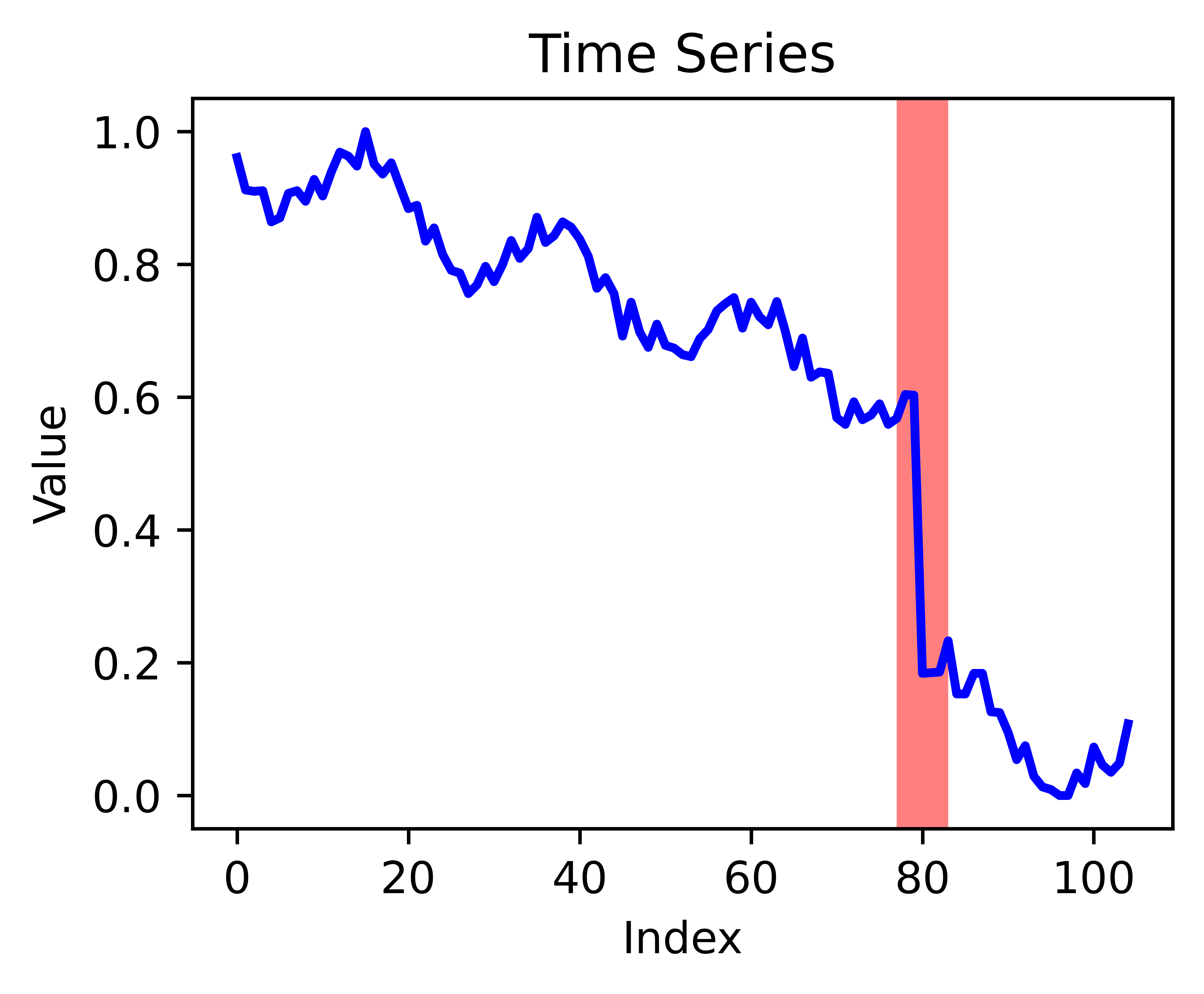}
    \caption{Line-Level}
  \end{subfigure}

  \vspace{5pt} % 行间距

  % 第2行
  \begin{subfigure}[b]{0.22\textwidth}
    \includegraphics[width=\linewidth]{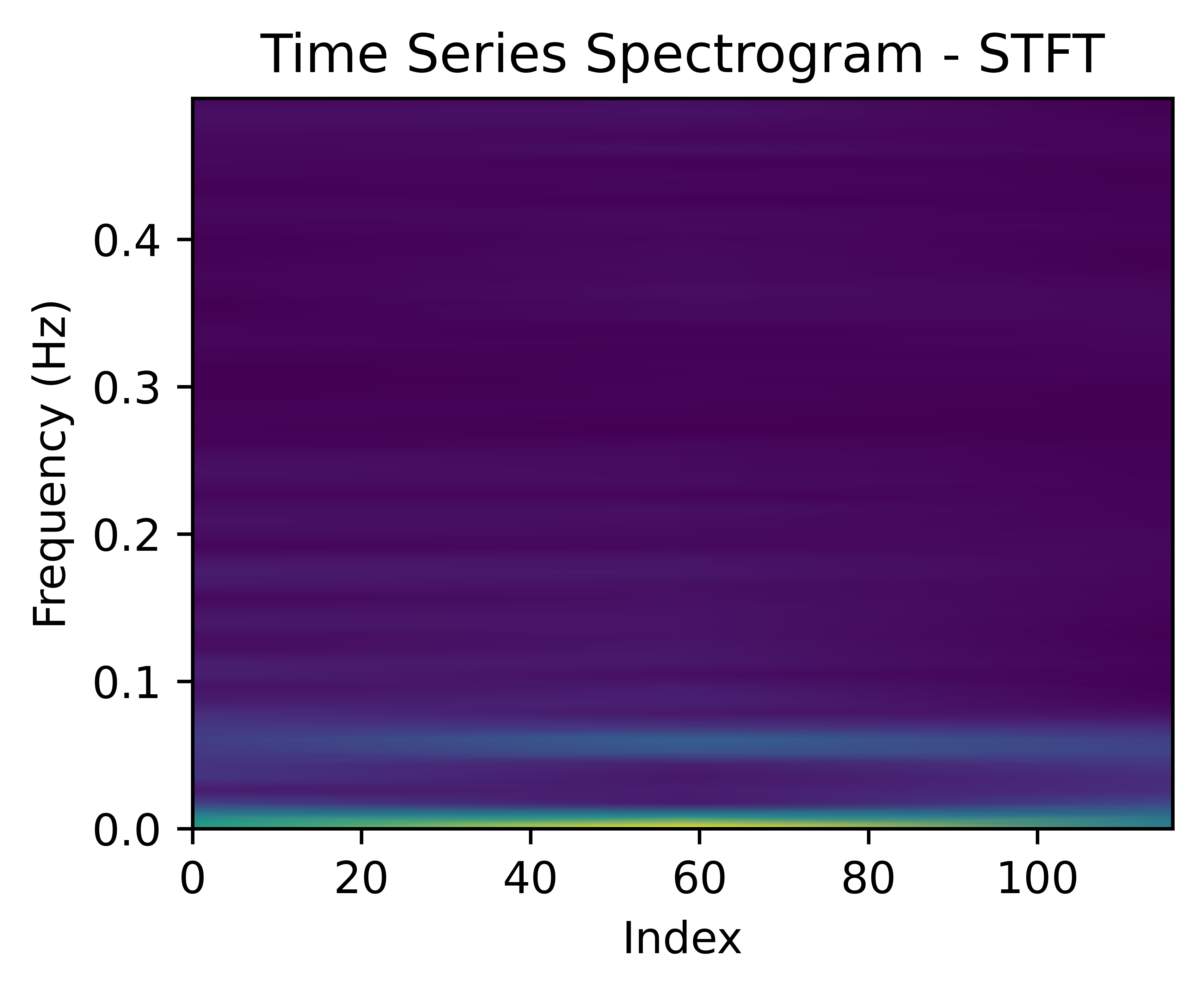}
    \caption{STFT-Spike}
  \end{subfigure}
  \hfill
  \begin{subfigure}[b]{0.22\textwidth}
    \includegraphics[width=\linewidth]{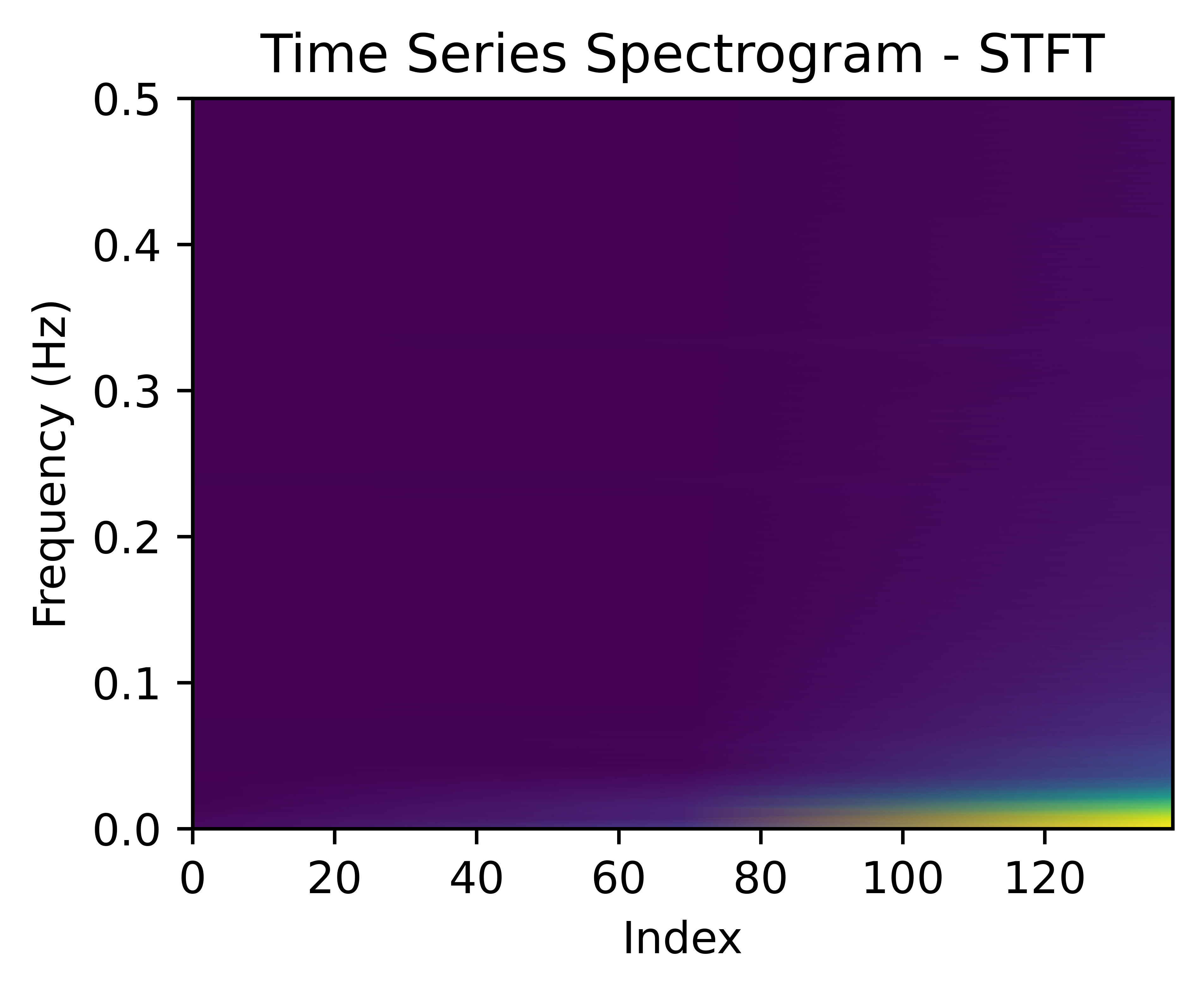}
    \caption{STFT-Trend}
  \end{subfigure}
  \hfill
  \begin{subfigure}[b]{0.22\textwidth}
    \includegraphics[width=\linewidth]{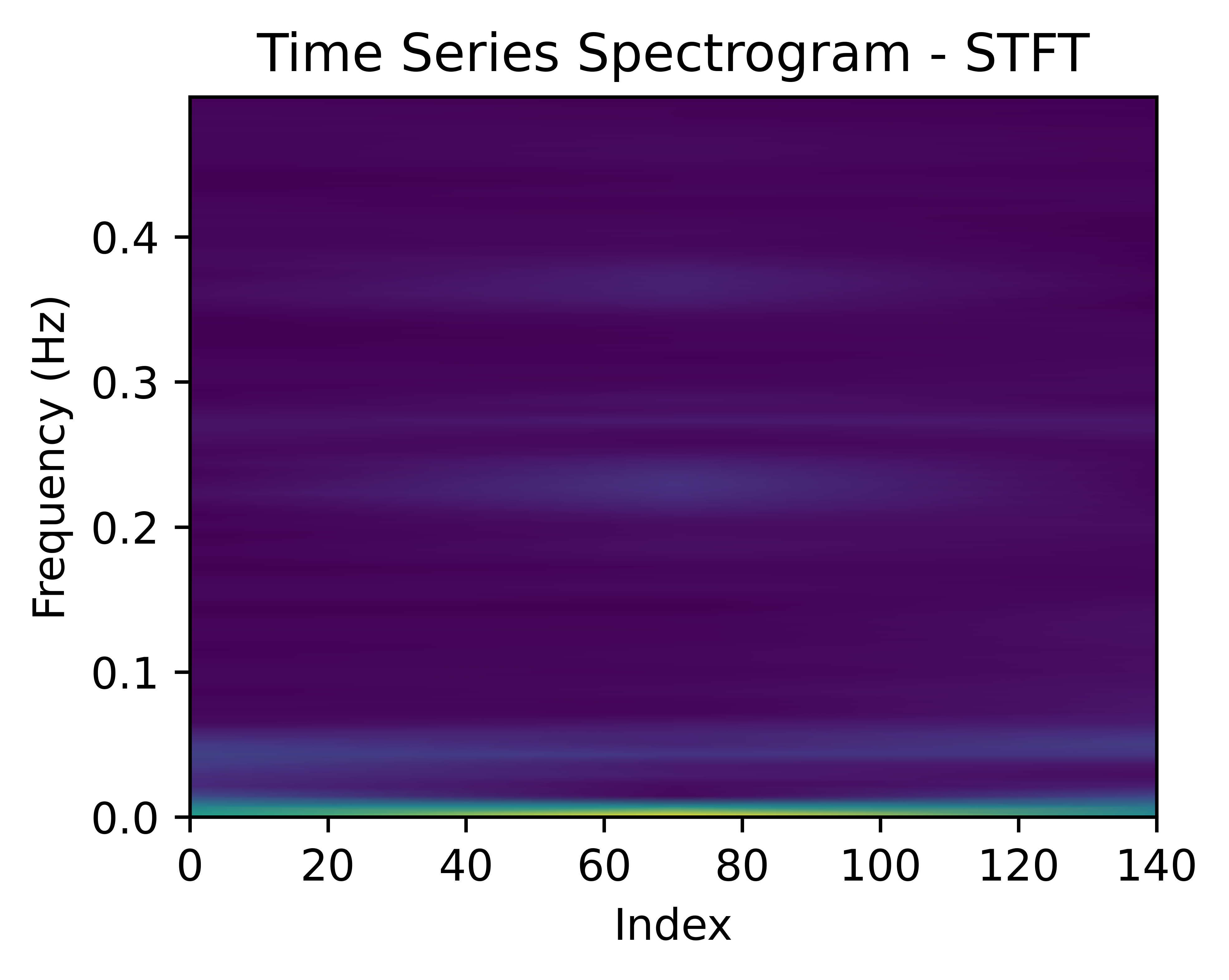}
    \caption{STFT-Frequency}
  \end{subfigure}
  \hfill
  \begin{subfigure}[b]{0.22\textwidth}
    \includegraphics[width=\linewidth]{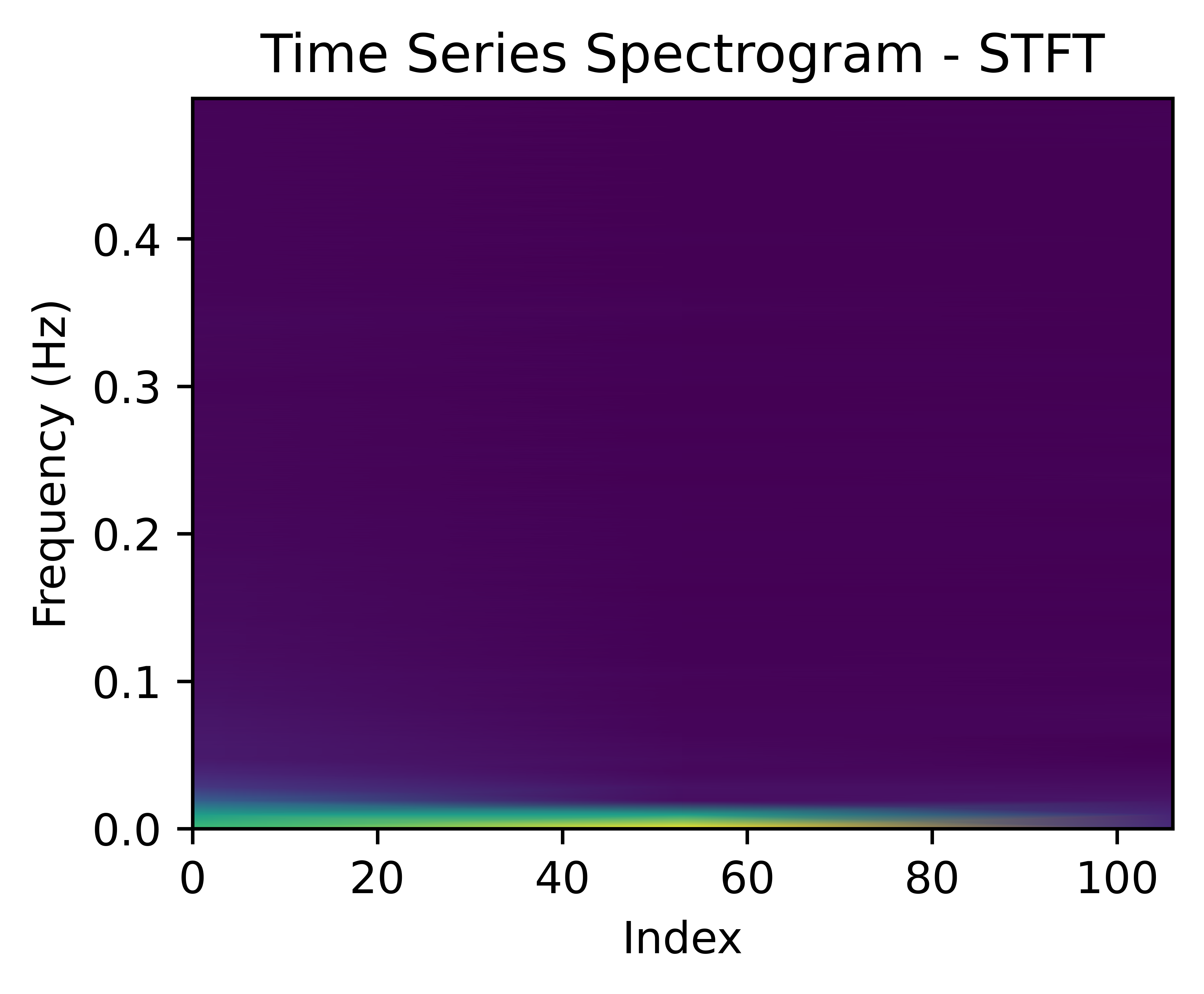}
    \caption{STFT-Level}
  \end{subfigure}

  \vspace{5pt}

  % 第3行
  \begin{subfigure}[b]{0.22\textwidth}
    \includegraphics[width=\linewidth]{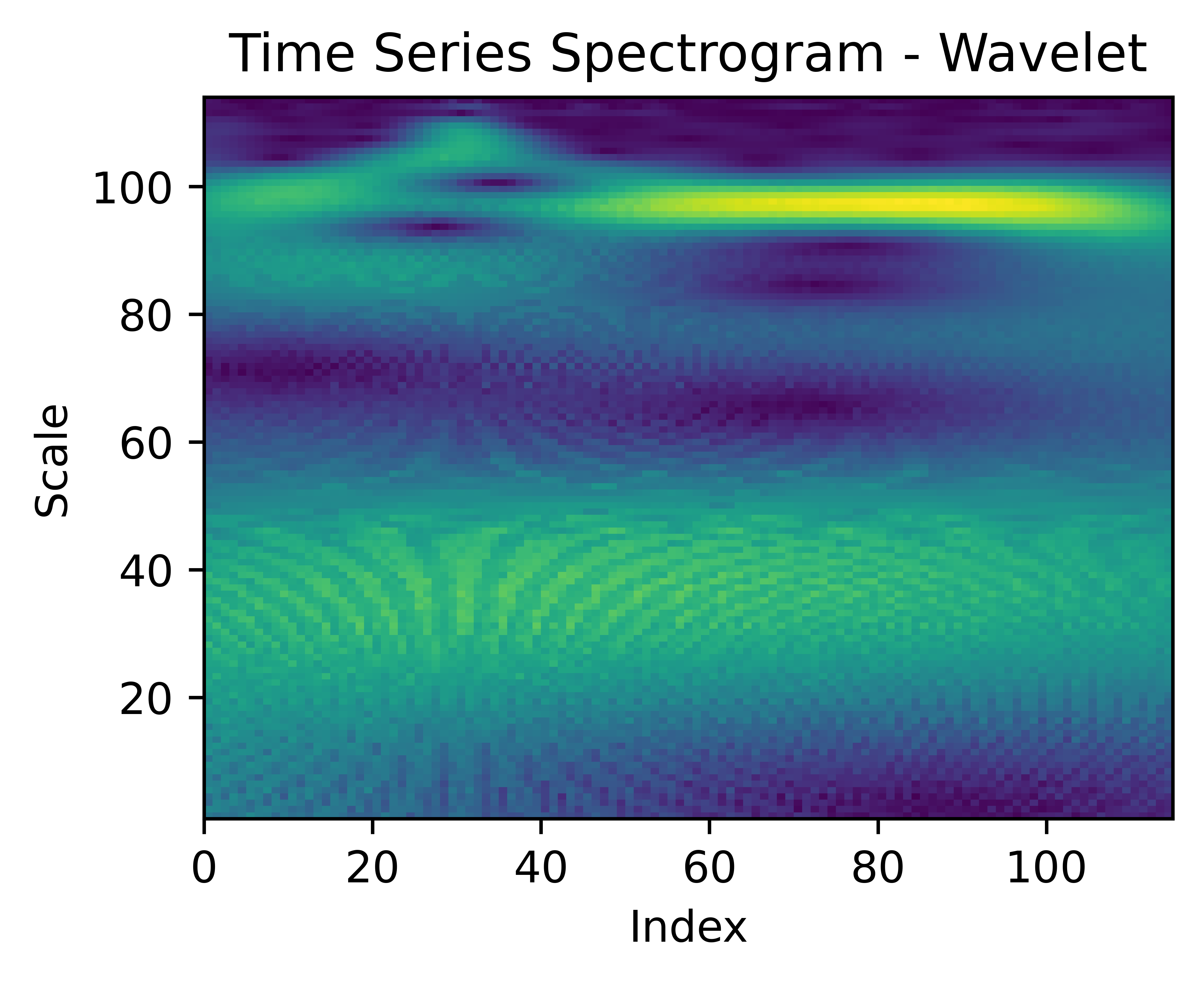}
    \caption{Wavelet-Spike}
  \end{subfigure}
  \hfill
  \begin{subfigure}[b]{0.22\textwidth}
    \includegraphics[width=\linewidth]{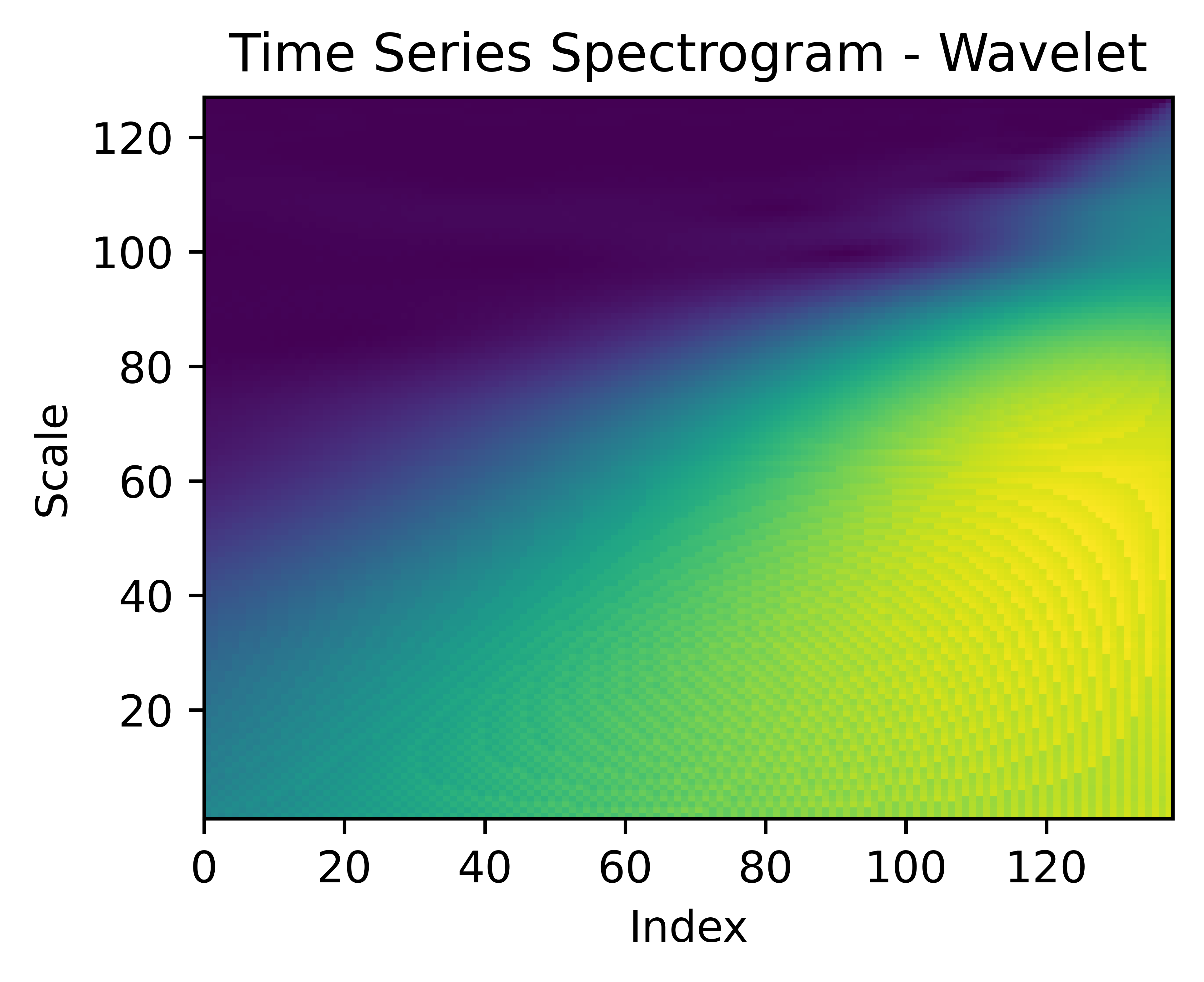}
    \caption{Wavelet-Trend}
  \end{subfigure}
  \hfill
  \begin{subfigure}[b]{0.22\textwidth}
    \includegraphics[width=\linewidth]{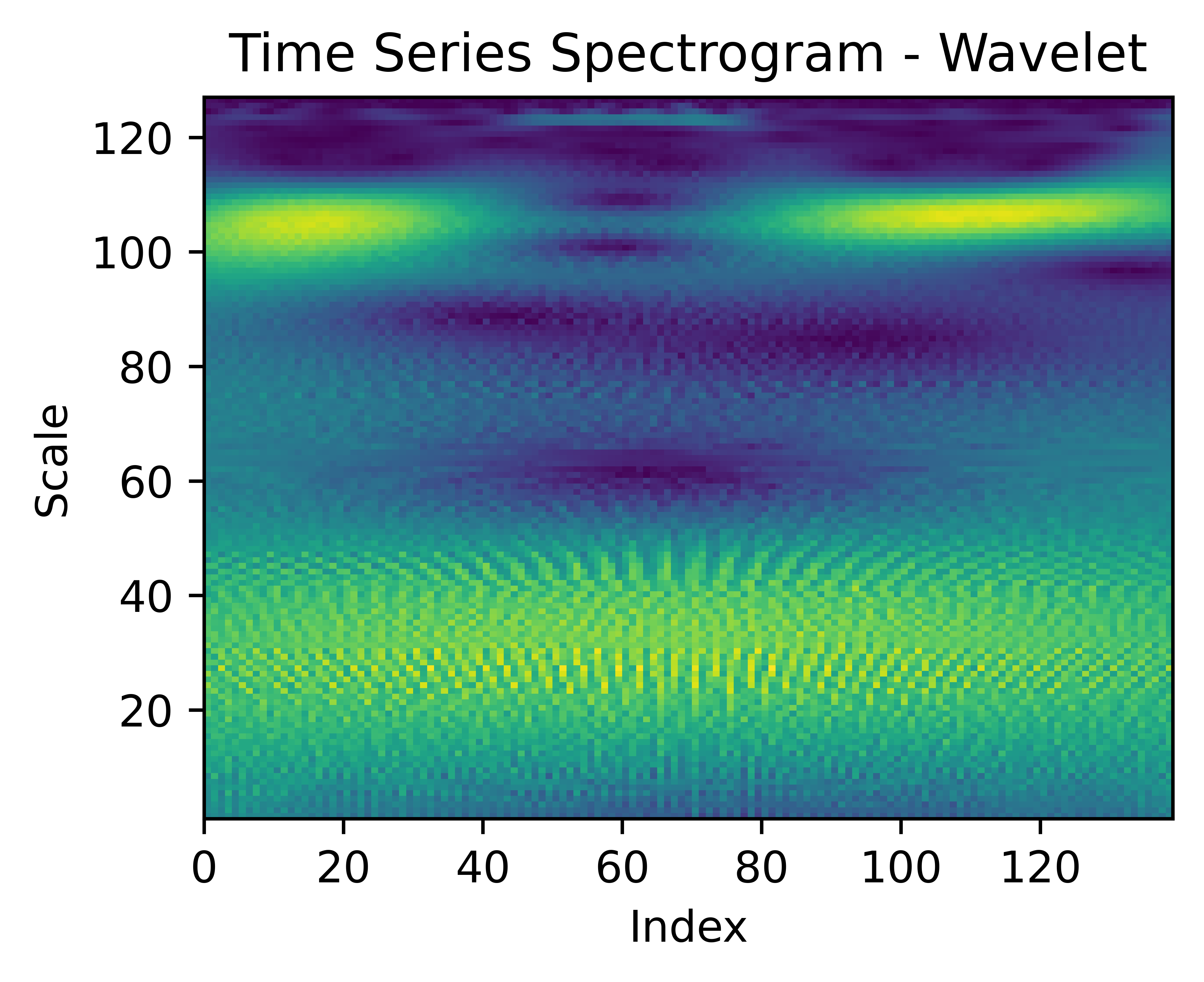}
    \caption{Wavelet-Frequency}
  \end{subfigure}
  \hfill
  \begin{subfigure}[b]{0.22\textwidth}
    \includegraphics[width=\linewidth]{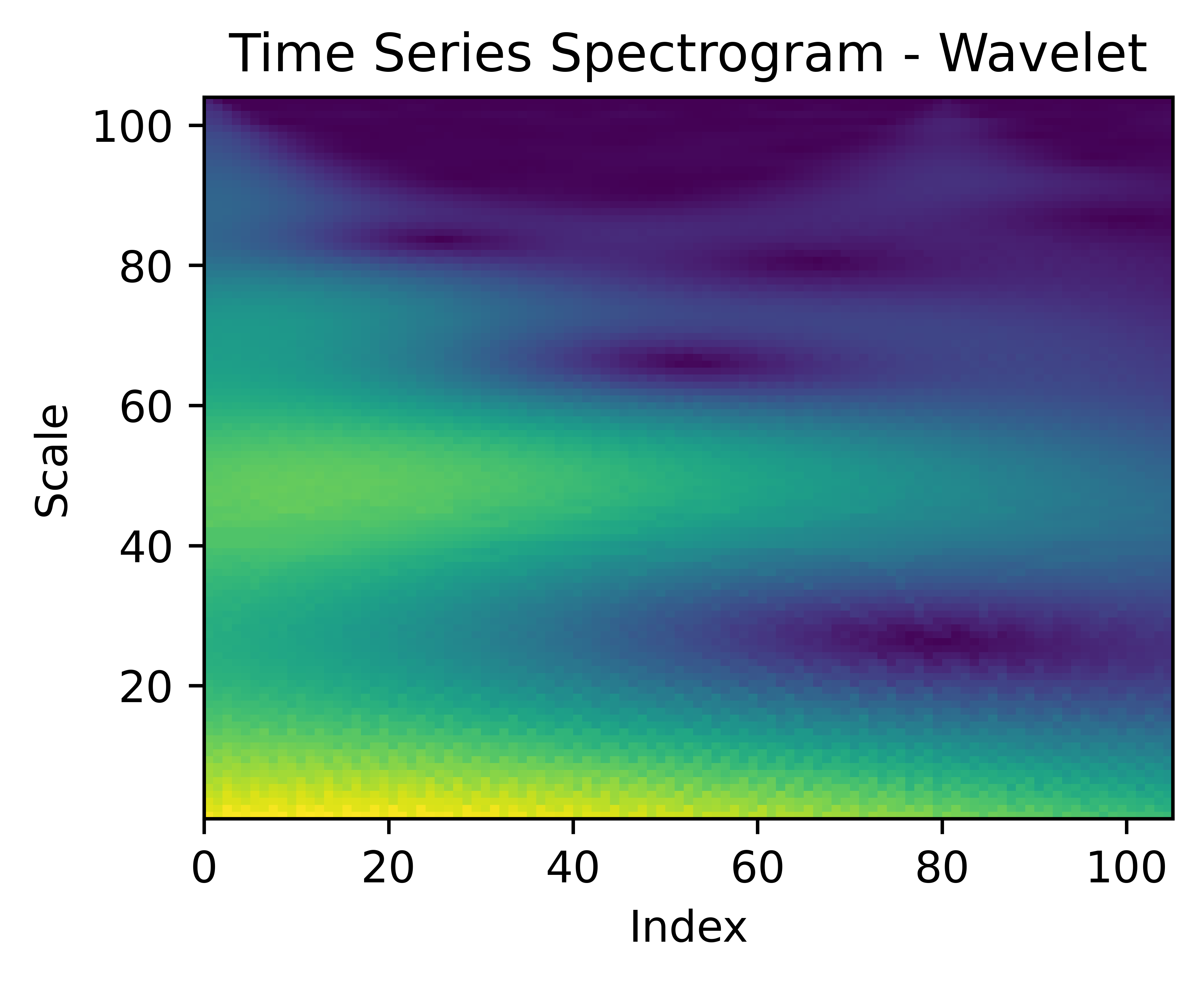}
    \caption{Wavelet-Level}
  \end{subfigure}

  \caption{LINE, STFT and Wavelet visualizations of four types of anomalies.}
  \label{fig:3x4grid}
\end{figure*}

Shown in Figure \ref{fig:3x4grid}.
\label{sec:visual}

\section{Output of Different Image Types}
\label{imagetype}

\noindent\fbox{%
  \parbox{\linewidth}{%
  \small
    \textbf{LINE:} \textit{Key observations from the chart:
    1. The series starts at a low value around 0.00 and gradually increases. 
    2. There are fluctuations in the series, indicating variability or noise in the data. 
    3. A significant peak occurs around the 125th index, reaching a value close to 1.00. 
    4. After the peak, the series fluctuates but generally decreases and stabilizes around a value of 0.50.}
    \\
    \textbf{LINE\_STFT:} \textit{The image consists of two plots. The top plot is labeled "Time Series (Index-based)" and shows a time series of data indexed from 0 to 160. The y-axis represents the "Value," which ranges from 0 to 1. The data appears to be a noisy signal with some peaks and troughs, particularly noticeable around index 125 where there is a sharp peak. The bottom plot is labeled "STFT Magnitude Spectrum."} 
  }%
}

\section{Experiment Setting}
\label{sec:expsetting}
% 以下简单介绍一些实验设置，更多的细节请参考代码。

% \subsection{Zero-shot Setting}

% zero-shot setting 指的是测试和训练数据不同。因此，在公开数据集上进行评测时，ViTs使用合成数据训练，即没有使用任何真实世界数据集训练。对于其他的SOTA方法，则使用除了评估数据集合以外的数据集进行训练。即评测KPI数据集，则使用Yahoo、WSD、NAB、UCR、TODS数据集训练。通过这种方式，对ViTs和其他方法进行相对公平的对比。

% \subsection{Training Data}

% 训练数据总计15k个QA，第一阶段5k个 时序描述的QA，第二个和第三个阶段均为相同的10k个TSAD QA。QA中涉及的time series 长度均为200.均使用文中提出的基于STL的方法生成。其中四类异常被随机的插入。需要注意的是，frequency和trend 异常只在较短周期的时序中插入，因为只有周期较短，周期数目足够长，才能观测出异常。

% \subsection{Testing Data}

% 测试数据包含两部分。合成测试数据使用文中的方法生成，共计2k个TSAD QA。公开数据集上，为了测试效率，我们选取Yahoo数据集每条时序的后50%，KPI、WSD数据集每条时序的后20%，作为测试集合。之后采用滑动窗口的方式生成不同的检测窗口。为了测试速度，我们对于KPI和WSD分别选取window=200，step=200，resize=1，对于Yahoo数据集，由于其每个点代表一个小时，周期较短，因此设置window=100，step=100，resize=0.5. 这里选择step=100的原因是，我们的方法已经够准确，每个点测试一次便可以满足要求。

% \subsection{Training Hyperparameters}

% 对于前两个SFT训练阶段，我们使用LLaMA-Factory框架进行全参数训练。每个阶段训练一个epoch，learning_rate设置为1.0e-5。RL阶段，我们使用EasyR1训练矿机，由于收敛较慢，我们训练两个epoch，learning rate设置为 1.0e-6。

The following briefly introduces some experimental settings. For more details, please refer to the code.

\subsection{Zero-shot Setting}

The zero-shot setting refers to the scenario where the test and training data are different. Therefore, when evaluating on public datasets, ViTs is trained using synthetic data, without using any real-world datasets. For other SOTA methods, training is conducted using datasets other than the evaluation datasets. For example, when evaluating the KPI dataset, training is done using the Yahoo, WSD, NAB, UCR, and TODS datasets. This allows for a relatively fair comparison between ViTs and other methods.

\subsection{Training Data}

The training data consists of a total of 15k QA pairs. The first stage includes 5k time series description QA pairs, while the second and third stages each include 10k TSAD QA pairs. The time series length in the QA pairs is consistently 200, generated using the STL-based method proposed in the paper. Four types of anomalies are randomly inserted. It is important to note that frequency and trend anomalies are only inserted into time series with shorter periods because only short periods with sufficient periodic instances can effectively display the anomalies.

\subsection{Testing Data}

The testing data includes two parts. The synthetic test data, generated using the method described in the paper, consists of 2k TSAD QA pairs. For public datasets, to ensure testing efficiency, we select the last 50\% of each time series from the Yahoo dataset and the last 20\% from the KPI and WSD datasets as the test sets. A sliding window approach is then employed to generate different detection windows. To optimize testing speed, we set the window size to 200, step size to 200, and resize factor to 1 for the KPI and WSD datasets. For the Yahoo dataset, due to the shorter duration represented by each data point (one hour), we set the window size to 100, step size to 100, and resize factor to 0.5. The step size is set to 100 because our method is accurate enough to require only one test per point.

\subsection{Pubulic Dataset Details}

% \begin{itemize}
\noindent\textbf{Yahoo} \cite{yahoo}
Yahoo is an open data set for anomaly detection released by Yahoo lab. The dataset consists of real and synthetic time series with tagged anomaly points. The dataset tests the detection accuracy of various anomaly-types including outliers and change-points. The synthetic dataset consists of time series with varying trend, noise and seasonality. The real dataset consists of time series representing the metrics of various Yahoo services.

\noindent\textbf{KPI} \cite{aiops}
KPI is collected from five large Internet companies (Sougo, eBay, Baidu, Tencent, and Ali). Specifically, this dataset includes 29 Curves in total. For each curve, experienced engineers in these companies
manually labeled anomalies carefully. Most KPI curves have an interval of 1 minute between two adjacent data points, while some of them have an interval of 5 minutes.

\noindent\textbf{WSD} \cite{wsd2}
Web service dataset (WSD) contains real-world KPIs collected from three top-tier Internet companies, Baidu, Sogou, and eBay, providing large-scale Web services. Experienced operation engineers have carefully labeled all KPIs in WSD.
% \end{itemize}

\subsection{Training Hyperparameters}

For the first two SFT training stages, we use the LLaMA-Factory framework for full-parameter training, with one epoch for each stage and a learning rate of 1.0e-5. During the RL stage, we use the EasyR1 training framework. Due to slower convergence, we train for two epochs with a learning rate of 1.0e-6.

\begin{table*}[t]
\centering
\caption{Performance of different settings on synthetic data.}
\label{tab:fullablation}
\resizebox{\textwidth}{!}{
\begin{tabular}{cccccccccccccccccccc} 
\toprule
\textbf{Setting}                &                   & \multicolumn{3}{c}{\textbf{Spike}} & \multicolumn{3}{c}{\textbf{Trend}} & \multicolumn{3}{c}{\textbf{Level Shift}} & \multicolumn{3}{c}{\textbf{Frequency}} & \multicolumn{3}{c}{\textbf{Mixed}} & \multicolumn{3}{c}{\textbf{Overall}}  \\ 
\cmidrule{3-20}
                                &                   & P & R      & F1                    & P      & R      & F1               & P      & R      & F1                     & P      & R      & F1                   & P       & R        & F1            & P      & R      & F1                  \\ 
\midrule
\textbf{Naive}                  &                   & 0.5631 & 0.5490 & 0.5560               & 0.7019 & 0.0913 & 0.1616          & 0.5733 & 0.5819 & 0.5776                 & 0.8993 & 0.3035 & 0.4538               & 0.6910  & 0.3107   & 0.4287        & 0.3813 & 0.4216 & 0.4004              \\
\textbf{\#1 SFT-\#2 SFT}        & \#1 Frozen LLM    & 0.9395 & 0.8222 & 0.8769               & 0.8420 & 0.0630 & 0.1172          & 0.9523 & 0.8714 & 0.9100                 & 0.9468 & 0.4909 & 0.6465               & 0.9225  & 0.5405   & 0.6816        & 0.9388 & 0.6280 & 0.7526              \\
\textbf{\#1 SFT-\#2 SFT}        & \#1 Full          & 0.9522 & 0.8104 & 0.8756               & 0.9019 & 0.2247 & 0.3597          & 0.9377 & 0.8574 & 0.8958                 & 0.9767 & 0.5454 & 0.6999               & 0.9115  & 0.5804   & 0.7092        & 0.9483 & 0.6602 & 0.7785              \\
\textbf{\#2 SFT}                &                   & 0.9347 & 0.8203 & 0.8737               & 0.9365 & 0.1543 & 0.2649          & 0.9459 & 0.8732 & 0.9081                 & 0.9328 & 0.4813 & 0.6350               & 0.9038  & 0.5038   & 0.6470        & 0.9345 & 0.6400 & 0.7597              \\
\textbf{\#3 RL}                 &                   & 0.7850 & 0.7330 & 0.7582               & 0.9187 & 0.6646 & 0.7713          & 0.6922 & 0.6481 & 0.6695                 & 0.9184 & 0.8154 & 0.8638               & 0.9025  & 0.5424   & 0.6776        & 0.5892 & 0.7101 & 0.6440              \\
\textbf{\#1 SFT-\#2 SFT-\#3 RL} & \#3 Full          & 0.9565 & 0.8835 & 0.9185               & 0.9009 & 0.6949 & 0.7846          & 0.9687 & 0.9404 & 0.9543                 & 0.9526 & 0.8336 & 0.8891               & 0.9516  & 0.6668   & 0.7842        & 0.8791 & 0.8380 & 0.8581              \\
\textbf{\#1 SFT-\#2 SFT-\#3 RL} & \#3 Frozen Vision & 0.9591 & 0.8895 & 0.9230               & 0.8822 & 0.6302 & 0.7352          & 0.9681 & 0.9399 & 0.9538                 & 0.9608 & 0.8105 & 0.8793               & 0.9552  & 0.6176   & 0.7502        & 0.8829 & 0.8224 & 0.8516              \\
\bottomrule
\end{tabular}
}
\end{table*}

\section{Proof}
% {\small
% % \[
% % f(t) = a_0 + \sum_{n=1}^{\infty} \left( a_n \cos\left( \frac{2\pi n t}{T} \right) + b_n \sin\left( \frac{2\pi n t}{T} \right) \right) \]

% \[
% f(t) \approx a_0 + \sum_{n=1}^{N} \left( a_n \cos\left( \frac{2\pi n t}{T} \right) + b_n \sin\left( \frac{2\pi n t}{T} \right) \right) \]

% }

We prove that if \(f\) is a \(2\pi\)--periodic real function of bounded variation on \([0,2\pi]\), then its Fourier partial sums
\[
S_N(x) \;=\; \frac{a_0}{2} + \sum_{n=1}^N \bigl(a_n \cos nx + b_n \sin nx\bigr)
\]
satisfy the uniform error estimate
\[
\bigl\lvert f(x) - S_N(x)\bigr\rvert \;\le\; \frac{V_{0}^{2\pi}(f)}{\pi\,N},
\]
where \(V_{0}^{2\pi}(f)\) is the total variation of \(f\) on one period.

Let \(f\) be a real-valued, \(2\pi\)--periodic function of bounded variation on \([0,2\pi]\), with total variation
\[
V_{0}^{2\pi}(f)
:= \sup_{0 = x_0 < x_1 < \cdots < x_m = 2\pi}
\sum_{j=1}^m \bigl|f(x_j) - f(x_{j-1})\bigr|.
\]
Its Fourier coefficients are
\[
a_n = \frac{1}{\pi}\int_{0}^{2\pi} f(t)\cos(nt)\,dt,
\]
\[
b_n = \frac{1}{\pi}\int_{0}^{2\pi} f(t)\sin(nt)\,dt,
\quad n=0,1,2,\dots.
\]
Integrating by parts for \(n\ge1\) gives
% \[
% a_n = \frac{1}{\pi}\int_{0}^{2\pi}f(t)\cos(nt)\,dt
% = \left.\frac{f(t)\sin(nt)}{\pi n}\right|_{0}^{2\pi}
%   - \frac{1}{\pi n}\int_{0}^{2\pi}f'(t)\sin(nt)\,dt.
% \]
\begin{align*}
a_n &= \frac{1}{\pi}\int_{0}^{2\pi}f(t)\cos(nt)\,dt\\
&= \left.\frac{f(t)\sin(nt)}{\pi n}\right|_{0}^{2\pi}
  - \frac{1}{\pi n}\int_{0}^{2\pi}f'(t)\sin(nt)\,dt.
\end{align*}
By periodicity \(f(0)=f(2\pi)\), so the boundary term vanishes.  Interpreting \(f'(t)\,dt\) as the signed measure \(df(t)\) and taking absolute values yields
\[
|a_n| \le \frac{1}{\pi n}\int_{0}^{2\pi}\bigl|df(t)\bigr|
= \frac{V_{0}^{2\pi}(f)}{\pi\,n},
\]
and similarly \(\lvert b_n\rvert\le V_{0}^{2\pi}(f)/(\pi n)\).

Define the remainder after \(N\) terms by
\[
R_N(x) = f(x) - S_N(x)
= \sum_{|k|>N} c_k\,e^{ikx},
\]
where \(c_k\) are the complex Fourier coefficients.  Then
% \[
% \bigl|R_N(x)\bigr|
% \le \sum_{|k|>N}\bigl|c_k\bigr|
% = \sum_{n=N+1}^\infty\bigl(|c_n|+|c_{-n}|\bigr)
% \le \sum_{n=N+1}^\infty\frac{|a_n|+|b_n|}{1} 
% \le \sum_{n=N+1}^\infty \frac{2\,V_{0}^{2\pi}(f)}{\pi\,n}
% = \frac{2\,V_{0}^{2\pi}(f)}{\pi}\sum_{n=N+1}^\infty\frac1n.
% \]

\begin{align*}
\bigl|R_N(x)\bigr|
&\le \sum_{|k|>N}\bigl|c_k\bigr| \\
&= \sum_{n=N+1}^\infty\bigl(|c_n|+|c_{-n}|\bigr) \\
&\le \sum_{n=N+1}^\infty\frac{|a_n|+|b_n|}{1} \\
&\le \sum_{n=N+1}^\infty \frac{2\,V_{0}^{2\pi}(f)}{\pi\,n} \\
&= \frac{2\,V_{0}^{2\pi}(f)}{\pi}\sum_{n=N+1}^\infty\frac1n.
\end{align*}
Using the standard tail estimate \(\sum_{n=N+1}^\infty 1/n \le 1/N\), we conclude
\[
\bigl|f(x)-S_N(x)\bigr|\le\frac{V_{0}^{2\pi}(f)}{\pi\,N}.
\]
Thus the \(N\)-term Fourier partial sum \(S_N(x)\) approximates \(f(x)\)
uniformly with error \(O(1/N)\), as claimed.
\label{proof}

\section{QA Examples}

\subsection{Time Series Description QA}

\noindent\fbox{%
  \parbox{\linewidth}{%
  \small
    \textbf{Q:} \textit{<image>Given the time series visualization, analyze the time series.}
    \\
    \textbf{A:} \textit{The time series shows periodicity: The amplitude of the periodic fluctuation between point 0 and point 200 is 77.3, the period of the fluctuation is 96. From the perspective of the slope, the overall trend is decreasing. The overall noise standard deviation is around 0.02, very small compared the overall change of the curve. The curve is overall smooth with almost no noise.} 
  }%
}

\subsection{TSAD QA}

\noindent\fbox{%
  \parbox{\linewidth}{%
  \small
    \textbf{Q:} \textit{<image>Given the time series visualization, is there any anomaly in the time series? Output the anomalous intervals. \\ Output Foarmat: boxed{[[start1, end1], [start2, end2], ...]}}
    \\
    \textbf{A:} \textit{Final Answer:  boxed{[]}} 
  }%
}
\label{sec:QAExample}

\section{GRPO}

\begin{figure}[tbp]
    \centering
    \includegraphics[width=0.5\textwidth]{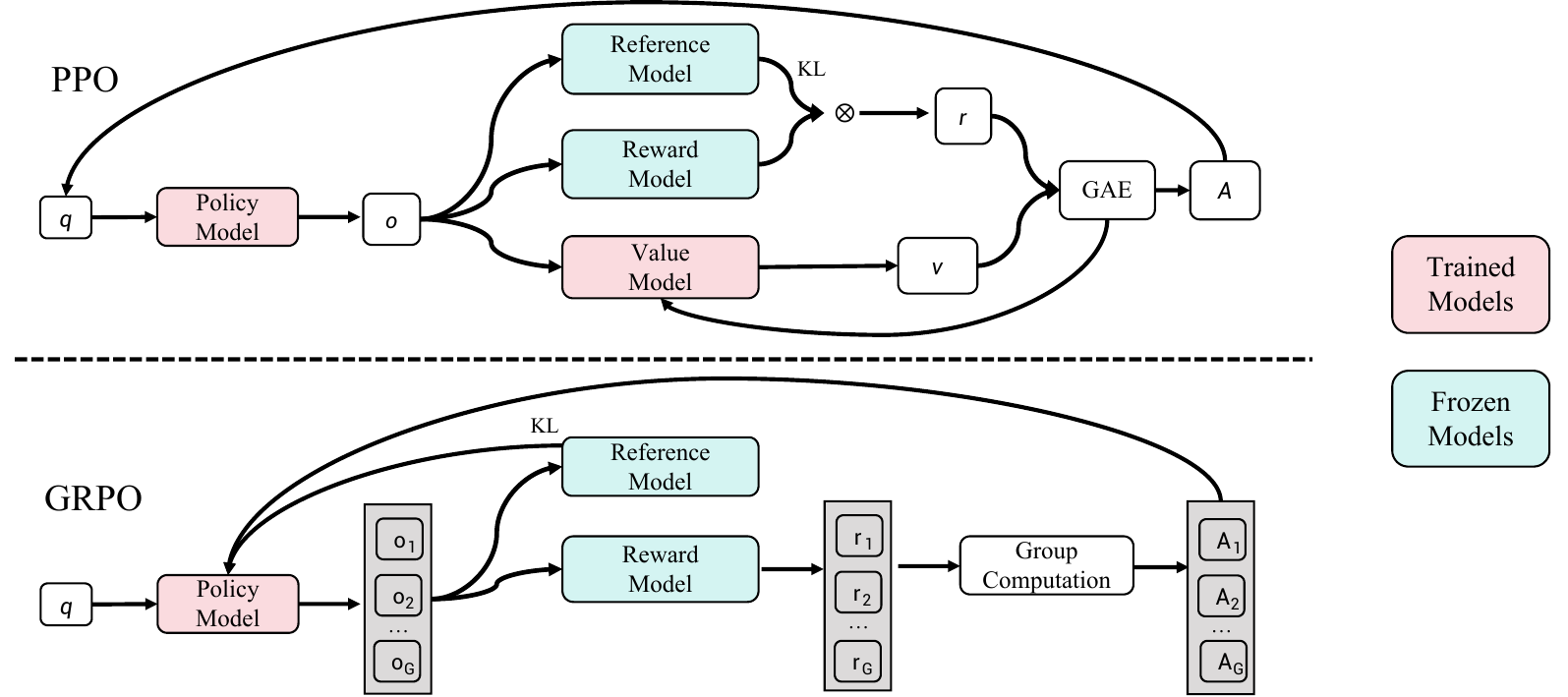} % 替换为你的文件名
    \caption{Camprision of GRPO and PPO.}
    \label{fig:grpo}
\end{figure}

% GRPO是目前相对有效的RL训练方式，相比传统的PPO需要需要训练一个价值函数来估计优势函数，GRPO的核心思想是通过组内相对奖励来估计基线（baseline），从而避免使用额外的价值函数模型（critic model）。具体来说，如图1所示，GRPO对于同一个输入，进行多次采样生成多个样本，之后根据预先定义好的reward model计算每个样本的reward。这里的reward model 一般是基于规则的计算方法，因为这样相对简单有效。得到一个group内所有的reward之后，如果说某个样本的reward大于group内部的平均reward则认为是优势样本。最终，GRPO通过不断迭代优化策略达到更优的效果。

GRPO is currently a relatively effective method for RL training. Unlike the traditional Proximal Policy Optimization (PPO), which requires the training of a value function to estimate the advantage function, the core idea of GRPO is to estimate the baseline using intra-group relative rewards. This approach eliminates the need for an additional value function model, or critic model. Specifically, as illustrated in Figure \ref{fig:grpo}, GRPO performs multiple samplings for the same input, generating multiple samples, which are then evaluated using a pre-defined reward model to compute the reward for each sample. Typically, this reward model employs rule-based computation methods, as they are relatively simple and effective. Once all the rewards within a group are obtained, any sample with a reward greater than the group’s average reward is considered an advantageous sample. Ultimately, GRPO iteratively optimizes the policy to achieve superior performance.

\label{sec:GRPO}

\section{Full Results}

Shown in Table \ref{tab:fullablation}.
\label{sec:full}

% \section{Reward}
% \label{sec:reward}

% % \begin{figure}[tbp]
% %     \centering
% %     \includegraphics[width=0.45\textwidth]{pic/val_f1_reward_line_plot.pdf} % 替换为你的文件名
% %     \caption{Reward during RL stage.}
% %     \label{fig:reward}
% % \end{figure}

% Shown in Figure \ref{fig:reward}.

% \bibliographystyle{acm} % 参考文献样式

\end{document}